\documentclass[10pt,twocolumn,letterpaper]{article}
\usepackage[pagenumbers]{cvpr} 
\usepackage{booktabs}       
\usepackage{amsfonts}       
\usepackage{nicefrac}       
\usepackage{microtype}      
\usepackage{xcolor}         
\usepackage{graphicx}
\usepackage{amsmath}
\usepackage{amssymb}
\usepackage{array}
\usepackage{multirow}
\usepackage{color}
\usepackage{colortbl}
\usepackage{framed}
\usepackage{bm}
\usepackage{xspace}
\usepackage{enumitem}
\usepackage{xparse}
\usepackage{algorithm}
\usepackage{listings}
\usepackage{url}
\usepackage{mathtools}
\usepackage{pifont}
\usepackage{mathtools}
\usepackage{lipsum}
\usepackage{adjustbox}
\usepackage{stfloats}
\usepackage{listings}

\newcommand{\cmark}{\ding{51}}%
\newcommand{\xmark}{\ding{55}}%

\definecolor{Gray}{gray}{0.2}
\definecolor{lightgray}{gray}{0.92}
\definecolor{OurColor}{rgb}{0.886, 0.941, 0.851}
\def \ie {\emph{i.e.}}
\def \eg {\emph{e.g.}}

\definecolor{customgray}{gray}{0.35}

\newcommand{\tit}[1]{\smallbreak\noindent\textbf{#1.}}
\newcommand{\tinytit}[1]{\noindent\textbf{#1.}}

\newcommand{\ours}{ReflectiVA\xspace}

\newcommand{\RET}{\texttt{\small{<RET>}}\xspace}
\newcommand{\NORET}{\texttt{\small{<NORET>}}\xspace}
\newcommand{\REL}{\texttt{\small{<REL>}}\xspace}
\newcommand{\NOREL}{\texttt{\small{<NOREL>}}\xspace}

\definecolor{backcolour}{rgb}{0.96,0.96,0.96}
\definecolor{GrayL}{gray}{0.3}
\definecolor{coolblack}{rgb}{0.0, 0.18, 0.39}
\definecolor{cornellred}{rgb}{0.7, 0.11, 0.11}
\definecolor{darktangerine}{rgb}{1.0, 0.66, 0.07}
\definecolor{deepcarrotorange}{rgb}{0.91, 0.41, 0.17}
\definecolor{noretcolor}{rgb}{0.847, 0.431, 0.8}
\definecolor{retcolor}{rgb}{0.627, 0.7686, 1.0}
\definecolor{relcolor}{RGB}{0, 176, 80}
\definecolor{imagecolor}{RGB}{254, 202, 203}
\definecolor{passagecolor}{RGB}{187, 232, 172}

\lstdefinestyle{mystyle}{
    backgroundcolor=\color{backcolour},   
    basicstyle=\ttfamily\footnotesize,
    breakatwhitespace=false,         
    breaklines=true,                 
    captionpos=b,                    
    keepspaces=true,                                 
    showspaces=false,                
    showstringspaces=false,
    showtabs=false,                  
    tabsize=2,
    escapeinside={(*}{*)},
}
\definecolor{cvprblue}{rgb}{0.21,0.49,0.74}
\usepackage[pagebackref,breaklinks,colorlinks,allcolors=cvprblue]{hyperref}

\newcommand\blfootnote[1]{%
  \begingroup
  \renewcommand\thefootnote{}\footnote{#1}%
  \addtocounter{footnote}{-1}%
  \endgroup
}

\def\paperID{****} 
\def\confName{CVPR}
\def\confYear{2025}

\title{Augmenting Multimodal LLMs with Self-Reflective Tokens for\\Knowledge-based Visual Question Answering}

\author{Federico Cocchi$^{*1,2}$, Nicholas Moratelli$^{*1}$, Marcella Cornia$^{1}$, Lorenzo Baraldi$^{1}$, Rita Cucchiara$^{1,3}$\\
$^1$University of Modena and Reggio Emilia, Italy\\
$^2$University of Pisa, Italy \quad $^3$IIT-CNR, Italy\\
{\tt\small \{name.surname\}@unimore.it}
}

\begin{document}
\maketitle

\begin{abstract}
Multimodal LLMs (MLLMs) are the natural extension of large language models to handle multimodal inputs, combining text and image data. They have recently garnered attention due to their capability to address complex tasks involving both modalities. However, their effectiveness is limited to the knowledge acquired during training, which restricts their practical utility. In this work, we introduce a novel method to enhance the adaptability of MLLMs by integrating external knowledge sources. Our proposed model, Reflective LLaVA (\ours), utilizes reflective tokens to dynamically determine the need for external knowledge and predict the relevance of information retrieved from an external database. Tokens are trained following a two-stage two-model training recipe. This ultimately enables the MLLM to manage external knowledge while preserving fluency and performance on tasks where external knowledge is not needed. Through our experiments, we demonstrate the efficacy of \ours for knowledge-based visual question answering, highlighting its superior performance compared to existing methods. Source code and trained models are publicly available at {\small\url{https://aimagelab.github.io/ReflectiVA}}.
\blfootnote{$^*$Equal contribution.}
\end{abstract}    
\section{Introduction}
\label{sec:intro}

\begin{figure}[t]
\vspace{-0.25cm}
    \centering
    \includegraphics[width=\linewidth]{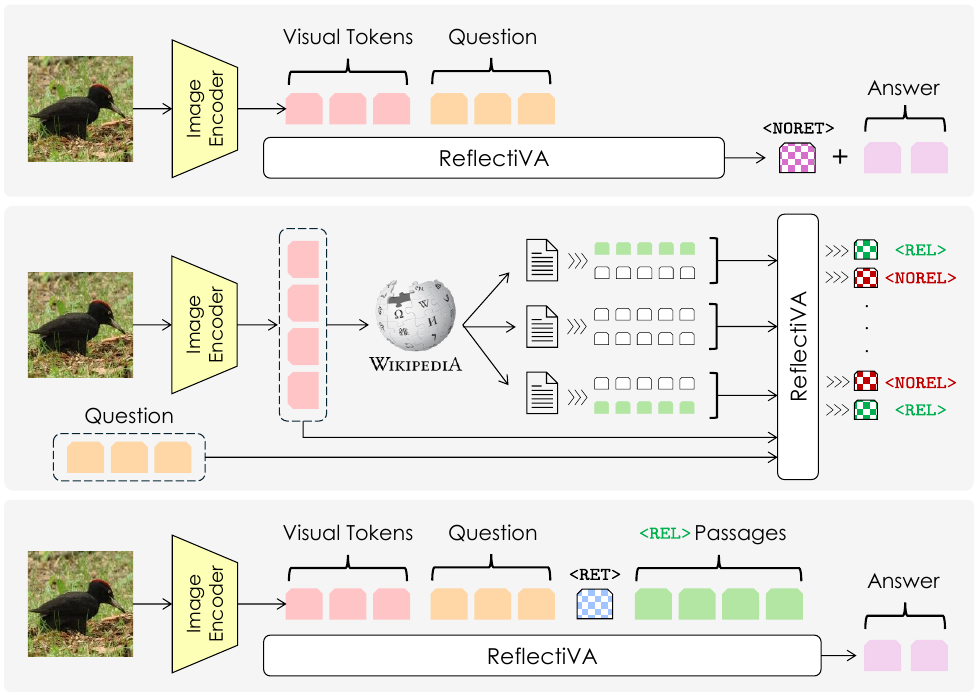}
    \vspace{-.6cm}
    \caption{Overview of \ours, which employs reflective tokens for knowledge-based visual question answering. Our model learns to predict the need of retrieving data from an external knowledge source (top), classifies the relevance of each retrieved item (middle) and generate the final answer based on relevant items (bottom).}
    \label{fig:first-page}
    \vspace{-.4cm}
\end{figure}

While the past few years have seen a surge of Large Language Models (LLMs) with increasing fluency and reasoning capabilities~\cite{brown2020language,raffel2020exploring,touvron2023llama}, thanks to the availability of large-scale training data and novel training techniques~\cite{chung2022scaling,ouyang2022training,jiang2024mixtral}, the Computer Vision community has recently started extending the capabilities of such models beyond pure text, with the inclusion of additional modalities like images, video, and 3D data. The resulting emergence of Multimodal Large Language Models (MLLMs) has been characterized by the development of models targeting multiple tasks~\cite{caffagni2024r,han2024onellm,sun2023generative} -- ranging from visual dialogue to image generation --, architectural innovations~\cite{li2023blip,alayrac2022flamingo,ye2024mplug}, and novel training recipes~\cite{liu2023improved,laurenccon2024obelics}.

What most existing MLLMs share, though, is their exclusive reliance on the knowledge learned at training time -- an issue that severely limits their practical applicability to cases that adhere to the training distribution. While this issue is also common to LLMs, it becomes more pressing in the case of MLLMs, where obtaining high-quality and large-scale multimodal data becomes even more difficult. Ideally, indeed, an MLLM should be capable of engaging in dialogues concerning specific visual details, long-tail knowledge, fine-grained categories, and instances~\cite{mensink2023encyclopedic,chen2023can}. However, this type of knowledge makes it hard for MLLMs to encode in their parameters because such long-tail information occurs rarely in the training data. Additionally, this lack of precise knowledge can also lead the MLLM to generate incorrect answers, thus again limiting their usage in practical cases.

A viable solution to this issue is to rely on a non-parametric approach, where content from external knowledge sources is incorporated into the MLLM context, conditioning the generation on accurate and relevant information~\cite{caffagni2024wiki,yan2024echosight}. However, building retrieval-augmented MLLMs presents unique challenges, ranging from the need to retrieve appropriate content, to the limited reasoning capabilities of MLLMs compared to LLMs. Further, an MLLM should also be capable of identifying \textit{when} external knowledge is needed, as opposed to answering questions that simply do not need external knowledge, \eg~purely visual questions. Lastly, retrieving relevant items from a multimodal knowledge source is an open problem, with state-of-the-art embedding spaces~\cite{radford2021learning} exhibiting limited performance~\cite{wei2023uniir,lin2024preflmr}. This further outlines the need to discover \textit{which} retrieved items may be relevant to answering a given user query.

Drawing inspiration from these challenges, in this paper, we propose a multimodal model for knowledge-based visual question answering which can jointly determine the need for accessing external knowledge and the relevance of items retrieved from an external knowledge base. Our model termed Reflective LLaVA (\ours), employs \textit{reflective tokens} to augment the capabilities of a pre-trained MLLM~\cite{liu2023improved} for knowledge-based generation. In particular, the vocabulary of the model is extended to generate additional tokens with which the model can decide whether retrieval is needed or not (Fig.~\ref{fig:first-page}, top), and if a retrieved sample is relevant or not for the input query (Fig.~\ref{fig:first-page}, middle). Training is conducted with a two-stage procedure employing two learnable models. We first train an in-article discriminator that can discriminate relevant passages from irrelevant ones found inside the same article. We then employ data annotated synthetically with this model, together with a mixture of existing datasets, to train the final model on all reflective tokens.

Experimentally, we evaluate the performance of the proposed approach on the Encyclopedic-VQA~\cite{mensink2023encyclopedic} and InfoSeek~\cite{chen2023can} datasets, which contain question-answer pairs linked with two knowledge bases derived from Wikipedia pages. Additionally, we assess the zero-shot generalization capabilities to two other VQA datasets~\cite{lerner2022viquae,jain2021select} which may require external knowledge to answer questions correctly. With extensive experiments, we demonstrate that our model outperforms previous works, and provides increased answer accuracy on all considered datasets and settings. Further, we demonstrate that the proposed approach maintains high performance on standard MLLM benchmarks~\cite{fu2023mme,yue2023mmmu,li2023evaluating} as well as on traditional VQA datasets~\cite{hudson2019gqa,singh2019towards} that do not require external knowledge during generation.
\section{Related Work}
\label{sec:related}

\tinytit{Multimodal Large Language Models}
Large Language Models (LLMs)~\cite{touvron2023llama,jiang2023mistral,dubey2024llama} have shown remarkable performance and adaptability across diverse tasks~\cite{vicuna2023,taori2023stanford}, with recent extensions incorporating multimodal capabilities, particularly integrating vision and language~\cite{mckinzie2024mm1,sun2023generative,li2024llava,caffagni2024r,cocchi2024llavamore}. A primary challenge in this domain is effectively combining LLMs with visual features and creating multimodal datasets for robust training. Architectural solutions for vision-to-language integration vary, including single linear projections or MLPs, as in the LLaVA models~\cite{liu2023visual,liu2023improved}, as well as Q-Former~\cite{li2023blip,dai2023instructblip} or Perceiver~\cite{laurenccon2024matters} modules to extract fixed-dimensional visual features. Models like Flamingo~\cite{alayrac2022flamingo,awadalla2023openflamingo} utilize cross-attention layers to directly integrate multimodal information. For training, multiple stages often leverage image captions for visual-text alignment~\cite{gadre2024datacomp,changpinyo2021conceptual}, and specialized datasets are developed for visual instruction tuning~\cite{liu2023visual,dai2023instructblip,laurenccon2024obelics}, with data quality and annotation specificity shown to significantly influence performance~\cite{deitke2024molmo}.

\tit{Knowledge-based Visual Question Answering}
This task involves answering questions that require external or specialized knowledge beyond the content of the image itself. Early datasets like OK-VQA~\cite{marino2019ok,schwenk2022okvqa} and KVQA~\cite{shah2019kvqa} introduced questions that require general and commonsense knowledge, which large-scale architectures such as MLLMs can increasingly handle within their current training scope. Newer datasets, such as Encyclopedic-VQA~\cite{mensink2023encyclopedic} and InfoSeek~\cite{chen2023can}, present greater challenges by focusing on highly specific, Wikipedia-scale knowledge. These require understanding detailed information about specific entities and nuanced encyclopedic facts. As a result, MLLMs often struggle in these settings, as they lack comprehensive coverage of this in-depth knowledge without relying on external sources.

To tackle this, contrastive image-text encoders~\cite{wei2023uniir,radford2021learning,xiao2024grounding,sun2024eva,lerner2024cross} are crucial for retrieving semantically aligned content based on image-question queries. Relevant passages from external knowledge sources are then accessed, with entities represented by both textual passages and images. Recent works~\cite{caffagni2024wiki,qi2024rora,yan2024echosight} have combined these retrieval methods with LLMs and MLLMs to enhance knowledge-based VQA tasks. For example, Wiki-LLaVA~\cite{caffagni2024wiki} integrates external multimodal knowledge via a hierarchical retrieval pipeline within a contrastive embedding space~\cite{radford2021learning}. RoRA-VLM~\cite{qi2024rora}, instead, introduces a visual token refinement to filter out query-irrelevant visual information from both retrieved and query images. Recently, inspired by advances in NLP~\cite{yu2024rankrag,ram2023context,asaiself}, EchoSight~\cite{yan2024echosight} proposes a Q-Former based re-ranking module to reorder retrieved textual passages before feeding them into the LLM. In contrast, our approach follows a different path enhancing MLLMs with specialized tokens that help to determine when retrieval is needed and assess the relevance of retrieved external knowledge.

\begin{figure*}[t]
    \centering
    \includegraphics[width=\textwidth]{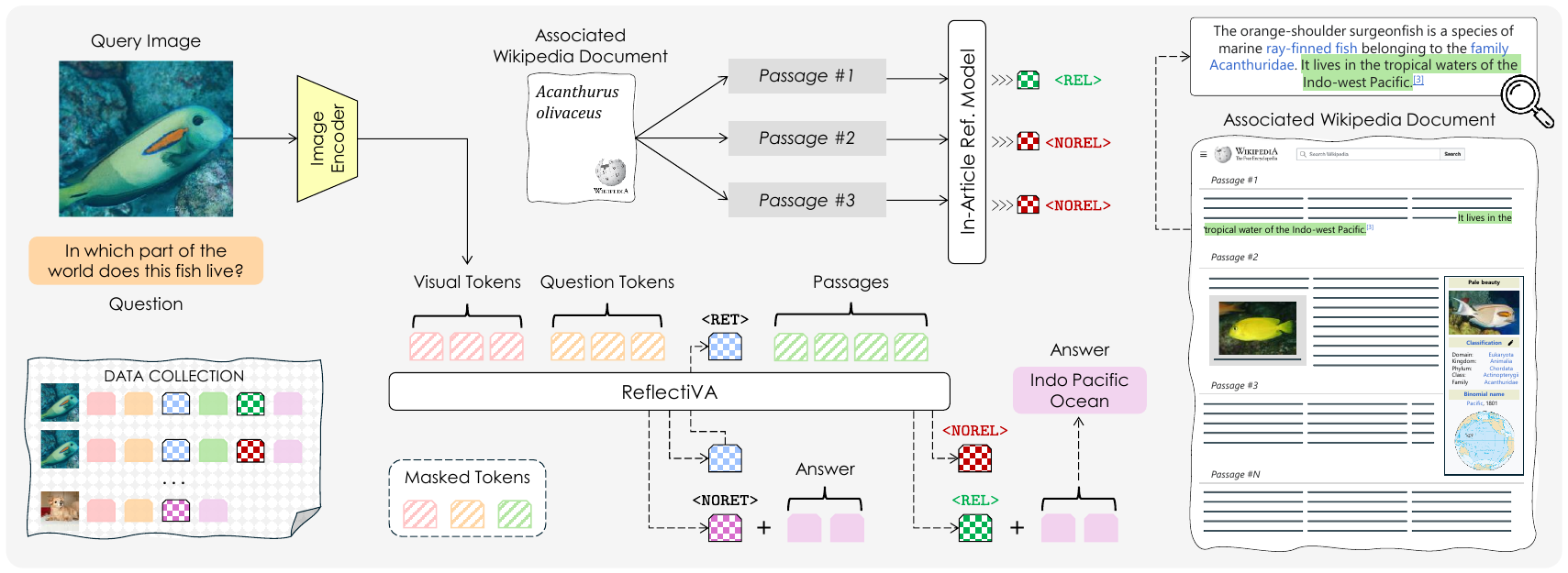}
    \vspace{-.6cm}
    \caption{Training approach for \ours. An in-article model is trained to predict the relevance of passages extracted from the ground-truth document corresponding to an $(I, q)$ pair. The in-article model then generates training data for \ours, which is trained to predict the need for external knowledge and the relevance of passages, along with the answer, using positive, soft- and hard-negative passages.}
    \label{fig:model}
    \vspace{-.35cm}
\end{figure*}

\tit{Retrieval-Augmented Language Models}
In recent years, retrieval-augmented generation has become increasingly popular in the field of LLMs, especially when training or fine-tuning models for specific tasks or domains is impractical. This has led to various techniques~\cite{guu2020retrieval,borgeaud2022improving,izacard2023atlas,jiang2023active,wang2024instructretro} for integrating additional information extracted from external sources to enhance the generation quality of frozen LLMs. Recent efforts in the NLP literature~\cite{asaiself} align with our proposal, introducing special tokens to improve generation and better incorporate retrieved data. To the best of our knowledge, however, we are the first to propose a token-guided retrieval-augmented generation pipeline in MLLMs, where multimodal understanding of both images and text is crucial.
\section{Proposed Method}
\label{sec:method}
\tinytit{Task Definition}
In retrieval-augmented generation, given a textual query $q$ and a query image $I$, an MLLM is expected to generate an answer $y$ by possibly leveraging additional snippets $\mathbf{S}$ retrieved from an external knowledge source as context. The objective of multimodal retrieval-augmented generation can therefore be written as
\begin{equation}
    y = \arg\max_y \text{MLLM}(y|I, q, \mathbf{S}).    
\end{equation}
In our setting, the external knowledge source $\mathcal{S}$ is composed of multimodal documents, each endowed with metadata (\eg, title and summary), textual passages organized in sections, and possibly visual content. Formally, the external database can be defined as a collection
\begin{equation}
    \mathcal{S} = \{  (\tilde{t}_i, \tilde{\mathbf{P}}_i, \tilde{I}_i) \}_{i=1}^N,
\end{equation}
where $\tilde{t}_i$ represents the metadata of a multimodal document, $\tilde{\mathbf{P}}_i$ the set of its textual passages, and $\tilde{I}_i$ its visual content.

\tit{Summary of the Approach}
To address the limitations of existing retrieval-augmented methods, our approach introduces two innovative strategies. Firstly, we enable the model to determine the optimal timing for retrieval, specifically when generating with an empty retrieval set $\mathbf{S}$ is advantageous because the query does not require external information. Secondly, after the retrieval process, we empower the model to identify the relevance of retrieved passages for generation. Both abilities are enabled through the incorporation of reflective tokens into the vocabulary of the model. These tokens are trained following a two-step two-model training recipe, which ultimately enables the MLLM with the ability to determine whether retrieval is needed and to select pertinent passages from the external database. An overview of our methodology is illustrated in Fig.~\ref{fig:model}.

\subsection{Adding Reflective Tokens}\label{sec:tokens}
Given a pre-trained MLLM, we augment its vocabulary $\mathcal{V}_0$ with a set of four additional tokens, \ie~$\{ \RET, \NORET, \REL, \NOREL \}$, which will enable the model to distinguish whether retrieval is needed (\RET, \NORET) and whether a retrieved sample is relevant to the input query (\REL, \NOREL). 

\tit{Generation Protocol}
At test time, the MLLM is prompted with an input image $I$ and a query $q$, and is asked to produce either the \RET or the \NORET token. If the \NORET token is sampled, the MLLM will be asked to directly generate the answer $y$ without relying on additional snippets. In this case, the generation process follows a schema
\begin{align}
    \texttt{\small{"<NORET>"}} &\sim \text{MLLM}\left([I, q], \{ \RET, \NORET \}, 1\right),\nonumber\\
    y &\sim \text{MLLM}([I, q, \NORET], \mathcal{V}_0),
\end{align}
where $y \sim \text{MLLM}(p, V, t)$ indicates that a sequence of tokens $y$ is sampled (\eg~through beam search) from the MLLM when prompted with an input $p$ and constrained to emit tokens belonging to a vocabulary $V$ and up to a length of $t$ tokens\footnote{When $t$ is omitted, we let the model generate until an ``end of sequence'' token is sampled.}. Finally, $[\cdot, \cdot]$ indicates concatenation. 

If instead the \RET token is sampled, we firstly retrieve a set of candidate textual passages $\mathbf{S}_0 = \{s_0, ..., s_k\}$ from the external knowledge base, and then ask the MLLM to evaluate the relevance of each of them through the emission of \REL and \NOREL tokens. In this case, the generation follows a protocol in the form
\begin{align}
    &\texttt{\small{"<RET>"}} \sim \text{MLLM}([I, q], \{ \RET, \NORET \}, 1), \\ \nonumber
    &r_i \sim \text{MLLM}([I, q, \RET, s_i], \{ \REL, \NOREL \}, 1),
\end{align}
where the second generation step is repeated for each item in $\mathbf{S}_0$ and $r_i$ indicates the relevance token sampled for each retrieved snippet.

After sampling relevance tokens, the MLLM is then asked to generate the final answer using the set of snippets that have been judged relevant, \ie~$\mathbf{S} = \{ s_i \in \mathbf{S}_0, r_i = \REL\}$. Formally, this generation stage is defined as
\begin{align}
    y &\sim \text{MLLM}([I, q, \RET, \mathbf{S}], \mathcal{V}_0),
\end{align}
where, for readability, the concatenation of all items in $\mathbf{S}$ is not made explicit.

\tit{Coarse-Grained Retrieval}
To obtain the set of candidate passages $\mathbf{S}_0$, we utilize the input query image $I$ as an anchor to retrieve a set of candidate documents from the knowledge source. We encode either the metadata or the image\footnote{Depending on the test case, see Sec.~\ref{sec:setup} for details.} of each document in the database using a CLIP-based textual or visual encoder~\cite{radford2021learning,sun2024eva} and build a dense vector-search database. The encoded features $\mathbf{z}_i \in \mathbb{R}^d$, where $d$ is the dimensionality of the CLIP embedding, act as a search index $\mathbf{Z} = \{ \mathbf{z}_i \}_i$. The retriever then employs a non-parametric function to compute the cosine similarity between the embedding of the query image and all search indexes. 
Based on this similarity search, the coarse-grained retriever retrieves the top-$k$ articles that are most similar to the query image. The set of candidate passages $\mathbf{S}_0$ is then built as the union of the passages belonging to the top-$k$ articles.

\subsection{Training an In-Article Reflective Model}
Clearly, a coarse-grained retrieval at the document level is not enough precise to retrieve the exact passage containing the answer. Indeed, this retrieval step is expected to have limited recall at lower values of $k$. Further, a more detailed examination of relevant documents is necessary to identify passages that can be utilized by the MLLM.

To this aim, we train the MLLM to emit the relevance tokens \REL and \NOREL. This is done with a two-stage, two-model training pipeline. Initially, we train an in-article reflective MLLM capable of distinguishing between relevant passages and negative passages from the same article. Subsequently, we employ predictions from that model to train the final MLLM with the ability to cope with negative passages taken from the same articles and from different articles.

\tit{Automatic Data Construction}
As most of the datasets for knowledge-based VQA do not provide human-labeled annotations of ground-truth textual passages, we employ a proprietary LLM to automatically annotate positive and negative passages. Given a query image $I$, a question $q$ and the set of passages from the associated article $\mathbf{P}$, we caption the image $I$ with three captioning models (\ie~LLaVA-v1.5~\cite{liu2023improved}, BLIP-2~\cite{li2023blip}, and InstructBLIP~\cite{dai2023instructblip}) and prompt the LLM to assess whether a passage $s \in \mathbf{P}$ can answer $q$ given the textual description of $I$. To help identify positive passages, for each sample, we select the two passages that have the highest similarity with their respective question according to the Contriever embedding space~\cite{izacard2021unsupervised}. We also ensure to have at least one positive and one negative passage for each $(I, q)$ sample in the dataset.

\tit{Model Training}
Having collected positive and negative passages, the in-article reflective model is trained on a mixture of samples associated with positive and negative passages, using sequences in the form
\begin{gather}
[I, q, \RET, s, \REL, y] \text{ and}\nonumber \\
[I, q, \RET, \tilde{s}, \NOREL, y], 
\end{gather}
where $s$ refers to a passage predicted as positive, and $\tilde{s}$ to a negative passage from the same page. The model is trained using a time-wise cross-entropy loss over the reflective tokens and over the answer.

\subsection{Training the Overall Model}
In the second stage, predictions from the in-article reflective model are employed to construct the dataset for training the overall model. 
The capabilities of the in-article model are indeed used to automatically annotate textual passages from existing datasets that require external knowledge. This is also complemented by negative passages taken from other pages of the knowledge base, plus samples that do not need an external knowledge base. The ultimate result of this stage is an MLLM capable of both answering questions and generating special tokens to assess whether additional information retrieval is necessary and whether it would be beneficial for answering.

\tit{Data Curation}
The training split of the Encyclopedic-VQA~\cite{mensink2023encyclopedic} and InfoSeek~\cite{chen2023can} datasets are employed to create the data collection to train the second stage. Each sample $(I, q)$ is expanded with three distinct passages: a positive, a hard negative coming from the ground-truth page, and a soft negative coming from a different page.

To construct the first two cases, each sample is processed by the in-article reflective model to label each passage as either relevant or not relevant. After this, the passage with the highest probability of containing the \REL token is taken as positive. Instead, the hard negative sample is randomly chosen from one of the sections predicted as \NOREL. For the soft negative case, the image $I$ is used to retrieve inside the top-1 page, excluding the ground-truth page. From the retrieved page, then, a random passage is considered a soft negative.
Additionally, data from the LLaVA-Instruct dataset~\cite{liu2023improved} are also included. These samples are labeled as cases where no retrieval is necessary.

\tit{Model Training}
Finally, the model is trained using a balanced mixture of sequences in the form
\begin{gather}
[I, q, \NORET, y], \nonumber\\
[I, q, \RET, s, \REL, y], \nonumber\\
[I, q, \RET, \tilde{s}, \NOREL, y] \text{ and} \nonumber\\ 
[I, q, \RET, \bar{s}, \NOREL, y], 
\end{gather}
where $\bar{s}$ refers to a soft-negative passage. We then employ a time-wise cross-entropy loss over all ground-truth tokens, with the exception of those related to $I$, $q$, and retrieved passages $s, \tilde{s}, \bar{s}$.
\section{Experiments}
\label{sec:experiments}

\subsection{Experimental Setup}
\label{sec:setup}

\tinytit{Datasets} 
Our experiments are conducted on Encyclopedic-VQA~\cite{mensink2023encyclopedic} and InfoSeek~\cite{chen2023can}, which contain question-answer pairs linked to documents from an external knowledge base (\eg~Wikipedia). Encyclopedic-VQA consists of 221k pairs, each paired with up to 5 images and associated with 16.7k fine-grained entities (\ie~Wikipedia pages). Questions are divided into single-hop and two-hop types: the former indicates that a single Wikipedia page is required to answer them, while the latter requires a sequential retrieval process across multiple documents. Dataset samples are split into training, validation, and test sets with 1M, 13.6k, and 5.8k items respectively. Experiments are reported on the test set, where single-hop questions correspond to 4.8k samples. The InfoSeek dataset, instead, contains 1.3M image-question pairs associated with around 11k Wikipedia pages. The dataset comprises 934k training, 73k validation, and 348k test samples. Following existing literature~\cite{caffagni2024wiki,yan2024echosight}, experimental results are reported on the validation set which includes questions not contained in the training split and questions associated with unseen entities.

\tit{External Knowledge Bases} Both datasets come with an external knowledge base composed of Wikipedia documents. In particular, Encyclopedic-VQA contains a knowledge base of 2M Wikipedia pages. Each page includes the Wikipedia title, the corresponding textual sections, and associated images. InfoSeek, instead, provides a knowledge base composed of 6M Wikipedia entities. In our experiments, we use the original 2M knowledge base for Encyclopedic-VQA, while we extract a subset of 100k pages\footnote{The knowledge base used for InfoSeek contains the same entities as~\cite{caffagni2024wiki}.} from the original 6M for InfoSeek, following recent works~\cite{caffagni2024wiki,yan2024echosight}.

\tit{Evaluation Metrics} We follow the evaluation protocol provided along with the datasets. Generated answers for Encyclopedic-VQA are evaluated according to the BERT matching score (BEM)~\cite{bulian2022tomayto} between predicted and ground-truth answers. Instead, when evaluating answers for image-question pairs from InfoSeek, we use VQA accuracy~\cite{goyal2017making} and relaxed accuracy~\cite{methani2020plotqa} depending on the question type.

\tit{Architectural and Training Details} Our model is based on the LLaVA-v1.5 MLLM~\cite{liu2023improved,cocchi2024llavamore} with LLaMA-3.1-8B~\cite{dubey2024llama} as language model\footnote{\scriptsize\url{https://huggingface.co/aimagelab/LLaVA_MORE-llama_3_1-8B-finetuning}}. It employs CLIP ViT-L/14@336 as visual encoder and an MLP as the vision-to-language connector. For both training phases, we fine-tune the LLaVA architecture to learn how to generate the introduced reflective tokens. To do that, we modify the original vocabulary of the LLaMA-3.1 LLM, replacing the final four reserved special tokens with our custom tokens. 
During training, we employ a learning rate of $2\times10^{-5}$ and a global batch size of $128$, updating the weights of both the MLP and LLM. The training is stopped based on the loss value on a separate validation split extracted from the training dataset.

\tit{Training Data Collection} 
We train our model on a collection of data from different sources. For both training phases, we include samples from the Encyclopedic-VQA and InfoSeek training splits, as well as data from LLaVA-Instruct~\cite{liu2023improved} to retain the generative capabilities of the MLLM. To automatically annotate positive and negative passages used to train the in-article reflective model, we employ GPT-4, prompted with the query question, the textual description of the associated image, the corresponding answer, and the passage to annotate. We also employ few-shot examples in the prompt to ease the task. Additional details on the training data mixture are reported in the supplementary material.

\begin{table}[t]
  \centering
  \setlength{\tabcolsep}{.2em}
  \resizebox{\linewidth}{!}{
  \begin{tabular}{lc c ccc c ccc}
   \toprule
    & & & \multicolumn{3}{c}{\textbf{E-VQA}} & & \multicolumn{3}{c}{\textbf{InfoSeek}} \\
    \cmidrule{4-6} \cmidrule{8-10}
    \textbf{Model} & \textbf{Ret. Mode} & & R@1 & R@5 & R@20 & & R@1 & R@5 & R@20 \\ 
    \midrule
    CLIP ViT-L/14 & Textual (T) & & 3.4 & 8.7 & 14.0 & & 36.9 & 59.9 & 71.9 \\ 
    CLIP ViT-L/14 & Textual (T+S) & & 0.7 & 2.3 & 6.4 & & 19.3 & 40.9 & 57.1 \\ 
    CLIP ViT-L/14 & Visual & & 9.9 & 22.0 & 31.7 & & 22.5 & 40.4 & 44.1 \\ 		
    \midrule 		
    EVA-CLIP-8B & Textual (T) & & 7.5 & 15.2 & 20.7 & & 52.5 & 71.2 & 79.2 \\ 
    EVA-CLIP-8B & Textual (T+S) & & 10.1 & 20.5 & 29.4 & & \textbf{56.1} & \textbf{77.6} & \textbf{86.4} \\ 
    EVA-CLIP-8B & Visual & & \textbf{15.6} & \textbf{36.1} & \textbf{49.8} & & 29.6 & 41.4 & 46.6 \\ 
  \bottomrule
  \end{tabular}
  }
  \vspace{-0.2cm}
\caption{Retrieval performance on Encyclopedic-VQA test set and the InfoSeek validation set. ``Textual'' refers to image-to-text retrieval using either title only (T) or title with summary (T+S), and ``Visual'' corresponds to image-to-image retrieval.}
  \label{tab:retrieval_small}
  \vspace{-0.35cm}
\end{table}

\begin{table*}[t]
  \centering
  \setlength{\tabcolsep}{.4em}
  \resizebox{0.97\linewidth}{!}{
  \begin{tabular}{lc c cc c cc c ccc}
   \toprule
    & & & & & & \multicolumn{2}{c}{\textbf{E-VQA}} & & \multicolumn{3}{c}{\textbf{InfoSeek}} \\
    \cmidrule{7-8} \cmidrule{10-12}
     \textbf{Model} & \textbf{LLM} & & \multicolumn{2}{c}{\textbf{Retrieval Mode}} & & Single-Hop & All & & Unseen-Q & Unseen-E & All \\
    \midrule
    \rowcolor{lightgray} 
    \multicolumn{12}{l}{\textit{Zero-shot LLMs}} \\
     Vanilla & Vicuna-7B & & \multicolumn{2}{c}{-} & & 2.1 & 2.0 & & 0.3 & 0.0 & 0.0 \\
      Vanilla & LLaMA-3-8B & & \multicolumn{2}{c}{-} & & 16.3 & 17.3 & & 1.5 & 0.0 & 0.0 \\
      Vanilla & LLaMA-3.1-8B & & \multicolumn{2}{c}{-}  & & 16.5 & 16.6 & & 2.1 & 0.0 & 0.0 \\
      Vanilla & GPT-4 & & \multicolumn{2}{c}{-} & & 21.9 & 23.4 & & 7.3 & 5.0 & 5.9 \\
    \midrule
    \rowcolor{lightgray} 
    \multicolumn{12}{l}{\textit{Zero-shot MLLMs}} \\
    BLIP-2~\cite{li2023blip} & Flan-T5$_\text{XL}$ & & \multicolumn{2}{c}{-} & &  12.6 & 12.4 & & 12.7 & 12.3 & 12.5 \\
    InstructBLIP~\cite{dai2023instructblip} & Flan-T5$_\text{XL}$ & & \multicolumn{2}{c}{-} & &  11.9 & 12.0 & & 8.9 & 7.4 & 8.1 \\
    LLaVA-v1.5~\cite{liu2023improved} & Vicuna-7B & & \multicolumn{2}{c}{-} & & 16.3 & 16.9 & & 9.6 & 9.4 & 9.5 \\
    LLaVA-v1.5~\cite{liu2023improved} & LLaMA-3.1-8B & & \multicolumn{2}{c}{-} & & 16.0 & 16.9 & & 8.3 & 8.9 & 7.8 \\
    GPT-4V~\cite{achiam2023gpt} & - & & \multicolumn{2}{c}{-} & & 26.9 & 28.1 & & 15.0 & 14.3 & 14.6 \\
    \midrule
    \rowcolor{lightgray} 
    \multicolumn{12}{l}{\textit{Retrieval-Augmented Models}} \\
    DPR$_\text{V+T}$~\cite{lerner2024cross}$^\dagger$ & Multi-passage BERT & & CLIP ViT-B/32 & Visual+Textual & & 29.1 & - & & - & - & 12.4 \\
    RORA-VLM~\cite{qi2024rora}$^\dagger$ & Vicuna-7B & & CLIP+Google Search & Visual+Textual & & - & 20.3 & & 25.1 & 27.3 & - \\
    Wiki-LLaVA~\cite{caffagni2024wiki} & Vicuna-7B & & CLIP ViT-L/14+Contriever & Textual & & 17.7 & 20.3 & & 30.1 & 27.8 & 28.9 \\
    Wiki-LLaVA~\cite{caffagni2024wiki}$^\diamondsuit$ & LLaMA-3.1-8B & & CLIP ViT-L/14+Contriever & Textual & & 18.3 & 19.6 & & 28.6 & 25.7 & 27.1 \\
    EchoSight~\cite{yan2024echosight}$^\dagger$ & Mistral-7B/LLaMA-3-8B & & EVA-CLIP-8B & Visual & & 19.4 & - & & - & - & 27.7 \\
    EchoSight~\cite{yan2024echosight}$^\diamondsuit$ & LLaMA-3.1-8B & & EVA-CLIP-8B & Textual & & 22.4 & 21.7 & & 30.0 & 30.7 & 30.4 \\
    EchoSight~\cite{yan2024echosight}$^\diamondsuit$ & LLaMA-3.1-8B & & EVA-CLIP-8B & Visual & & 26.4 & 24.9 & & 18.0 & 19.8 & 18.8 \\
    \rowcolor{OurColor}
    \textbf{\ours (Ours)} & LLaMA-3.1-8B & & CLIP ViT-L/14 & Textual & & 24.9 & 26.7 & & 34.5 & 32.9 & 33.7 \\
    \rowcolor{OurColor}
    \textbf{\ours (Ours)} & LLaMA-3.1-8B & & EVA-CLIP-8B & Textual & & 28.0 & 29.2 & & \textbf{40.4} & \textbf{39.8} & \textbf{40.1} \\
    \rowcolor{OurColor}
    \textbf{\ours (Ours)} & LLaMA-3.1-8B & & EVA-CLIP-8B & Visual & & \textbf{35.5} & \textbf{35.5} & & 28.6 & 28.1 & 28.3 \\
  \bottomrule
  \end{tabular}
  }
  \vspace{-0.2cm}
\caption{VQA accuracy scores on the Encyclopedic-VQA test set and the InfoSeek validation set, where all results from retrieval-augmented models are reported without considering any re-ranking stage to reorder retrieved documents. $\dagger$ indicates results that are not directly comparable due to different knowledge bases, and the marker $\diamondsuit$ represents our reproductions with different LLMs.
}
\label{tab:results}
\vspace{-0.35cm}
\end{table*}

\tit{Coarse-Grained Retrieval Details} To identify the set of documents most relevant to the query image and associated question, we evaluate two CLIP-based retrieval models, namely CLIP ViT-L/14@336~\cite{radford2021learning} and EVA-CLIP-8B~\cite{sun2024eva}. We explore two retrieval configurations for both models: (i) image-to-text retrieval, which computes similarity between the query image and document metadata (either the title alone or the title with the summary of the page), and (ii) image-to-image retrieval, which assesses similarity between the query image and images within Wikipedia documents. 

Retrieval results for each variant are detailed in Table~\ref{tab:retrieval_small}. Notably, EVA-CLIP demonstrates superior results across all configurations. However, the optimal retrieval configuration varies between datasets. Specifically, image-to-image retrieval yields the highest accuracy on Encyclopedic-VQA, while image-to-text retrieval proves the most effective for InfoSeek. This is probably due to the distinct characteristics and structural composition of each dataset and their respective knowledge bases. Consequently, unless specified otherwise, we adopt image-to-image retrieval for Encyclopedic-VQA and image-to-text retrieval (using title-only for CLIP ViT-L and title with summary for EVA-CLIP) for InfoSeek, with the number $k$ of retrieved documents equal to $5$.

\subsection{Comparison with the State of the Art}
\tinytit{Results on Encyclopedic-VQA and InfoSeek} 
We evaluate our model on the aforementioned datasets, comparing it to various zero-shot LLMs, MLLMs, and retrieval-augmented competitors. Specifically, we report results of four LLMs -- Vicuna~\cite{vicuna2023}, LLaMA-3, LLaMA-3.1~\cite{dubey2024llama} and GPT-4~\cite{achiam2023gpt} -- each prompted with both the query question and a description of the query image generated by an image captioning model. Additionally, we assess the performance of BLIP-2~\cite{li2023blip}, InstructBLIP~\cite{dai2023instructblip}, LLaVA-v1.5~\cite{liu2023improved} and GPT-4V~\cite{achiam2023gpt} without external retrieval augmentation and using only the query image and question as input. As direct competitors, we include  DPR~\cite{lerner2024cross}, RORA-VLM~\cite{qi2024rora}, Wiki-LLaVA~\cite{caffagni2024wiki}, and EchoSight~\cite{yan2024echosight}, which all leverage external knowledge retrieval. To ensure a fair comparison with the considered methods, for both Wiki-LLaVA and EchoSight we develop a variant based on LLaMA-3.1, employing the same knowledge bases used in our solution.

\begin{table}[t]
  \centering
  \setlength{\tabcolsep}{.45em}
  \resizebox{0.95\linewidth}{!}{
  \begin{tabular}{lc c cc}
   \toprule
    & & & \multicolumn{2}{c}{\textbf{E-VQA}} \\
    \cmidrule{4-5}
     \textbf{Model} & \textbf{LLM} & & Single-Hop & All \\
    \midrule
    \rowcolor{lightgray} 
    \multicolumn{5}{l}{\textit{Textual Retrieval Mode}} \\
     EchoSight~\cite{yan2024echosight}$^\diamondsuit$ & LLaMA-3.1-8B & & 26.8 & 26.0 \\
      \rowcolor{OurColor}
    \textbf{\ours (Ours)} & LLaMA-3.1-8B & & \textbf{33.6} & \textbf{33.9} \\
    \midrule
    \rowcolor{lightgray} 
    \multicolumn{5}{l}{\textit{Visual Retrieval Mode}} \\
    EchoSight~\cite{yan2024echosight}$^\dagger$ & Mistral-7B/LLaMA-3-8B & & \textbf{41.8} & - \\
     EchoSight~\cite{yan2024echosight}$^\diamondsuit$ & LLaMA-3.1-8B & & 36.3 & 34.2 \\
    \rowcolor{OurColor}
    \textbf{\ours (Ours)} & LLaMA-3.1-8B & & 40.6 & \textbf{39.7} \\
  \bottomrule
  \end{tabular}
  }
  \vspace{-0.2cm}
\caption{VQA accuracy scores on Encyclopedic-VQA when models are equipped with a document re-ranking component. $\dagger$ indicates results that are not directly comparable due to different knowledge bases, and the marker $\diamondsuit$ represents our reproductions.}
  \label{tab:results_reranking}
\vspace{-0.35cm}
\end{table}

Results are shown in Table~\ref{tab:results}, in which we report for each retrieval-augmented model the details of the retrieval pipeline used (\ie~the retrieval model and modality). As it can be seen, both zero-shot LLMs and MLLMs fail to correctly answer the given questions due to the lack of external knowledge during the generation. This is particularly evident on the InfoSeek dataset where LLMs exhibit accuracy scores close to zero, highlighting the need for the visual inputs to generate correct answers for this dataset. For knowledge-based models, the proposed \ours exhibits state-of-the-art results on both the Encyclopedic-VQA and InfoSeek datasets, outperforming all evaluated competitors by a substantial margin and highlighting the benefits of employing reflective tokens for the task.

\begin{table}[t]
  \centering
  \setlength{\tabcolsep}{.25em}
  \resizebox{0.97\linewidth}{!}{
  \begin{tabular}{lcc c c ccc}
   \toprule
    & & & \multicolumn{1}{c}{\textbf{E-VQA}} & & \multicolumn{3}{c}{\textbf{InfoSeek}} \\
    \cmidrule{4-4} \cmidrule{6-8}
    \textbf{Model} & \textbf{LLM} & & Single-Hop & & Un-Q & Un-E & All \\
    \midrule
    \rowcolor{lightgray} 
    \multicolumn{8}{l}{\textit{KB Article}} \\
    Vanilla & Vicuna-7B & & 34.1 & & 5.3 & 4.3 & 4.7 \\
    Vanilla & LLaMA-3-8B & & 72.9 & & 10.0 & 7.9 & 8.8 \\
    Vanilla & LLaMA-3.1-8B & & 73.6 & & 15.2 & 13.9 & 14.5 \\
    LLaVA-v1.5~\cite{liu2023improved} & Vicuna-7B & & 42.9 & & 14.2 & 13.4 & 13.8 \\
    LLaVA-v1.5~\cite{liu2023improved} & LLaMA-3.1-8B & & 54.1 & & 20.1 & 17.7 & 18.8 \\
    \midrule
    \rowcolor{lightgray} 
    \multicolumn{8}{l}{\textit{KB Passages}} \\
    Wiki-LLaVA~\cite{caffagni2024wiki} & Vicuna-7B & & 38.5 & & 52.7 & 50.3 & 51.5 \\
    Wiki-LLaVA~\cite{caffagni2024wiki}$^\diamondsuit$ & LLaMA-3.1-8B & & 46.8 & & 51.2 & 50.6 & 50.9 \\
    \rowcolor{OurColor}
     \rowcolor{OurColor}
    \textbf{\ours (Ours)} & LLaMA-3.1-8B & & \textbf{75.2} & & \textbf{57.8} & \textbf{57.4} & \textbf{57.6} \\
  \bottomrule
  \end{tabular}
  }
  \vspace{-0.2cm}
\caption{VQA accuracy scores on Encyclopedic-VQA and InfoSeek with oracle Wikipedia entities. The top of the table shows results using the full Wikipedia article as input to the LLM/MLLM, while the bottom of the table shows performance using model-specific strategies to identify relevant text passages.}
\label{tab:results_oracle}
\vspace{-0.35cm}
\end{table}

\tit{Integrating a Re-Ranking Stage} Recent studies focusing on text-only LLMs~\cite{ram2023context,yu2024rankrag,yan2024echosight} have shown that incorporating a re-ranking stage within a retrieval-augmented generation pipeline can enhance performance. Following this approach, we evaluate our model with the re-ranking component proposed in~\cite{yan2024echosight} which reorders retrieved textual passages prior to model input. Specifically, in our experiments, we first retrieve the top-$k$ relevant Wikipedia pages from the knowledge base using the retrieval model previously described. The retrieved passages are then processed by the re-ranking component, and we input the top-$k_p$ re-ranked passages into our model to assess their relevance to the query and generate a response\footnote{A detailed analysis of $k$ and $k_p$ in re-ranking settings is provided in the supplementary material.}. 

Results are shown in Table~\ref{tab:results_reranking} comparing our results to those of EchoSight on the Encyclopedic-VQA dataset\footnote{For this setting, we do not include results on InfoSeek since the re-ranker proposed in~\cite{yan2024echosight} is trained on samples from Encyclopedic-VQA.}. As it can be seen, performing a re-ranking step can further improve the results of our model which achieves, in its best configuration, 40.6 accuracy points compared to 35.5 without re-ranking. These results consistently outperform those obtained by EchoSigh, with the same LLM and knowledge base used in our setting, in both retrieval configurations.

\begin{table}[t]
  \centering
  \setlength{\tabcolsep}{.2em}
  \resizebox{\linewidth}{!}{
  \begin{tabular}{lc c c c ccc}
   \toprule
    & & & \multicolumn{1}{c}{\textbf{E-VQA}} & & \multicolumn{3}{c}{\textbf{InfoSeek}} \\
    \cmidrule{4-4} \cmidrule{6-8}
    & $k$ & & Single-Hop & & Un-Q & Un-E & All \\
    \midrule
    \rowcolor{lightgray}
    \multicolumn{8}{l}{\textit{Effectiveness of Reflective Tokens and Training}} \\
    In-Article Reflective Model & 5 & & 21.1 & & 25.5 & 23.8 & 24.6 \\
    Single model (w/ special tokens from LLaMA-3.1) & 5 & & 30.7 & & 27.3 & 28.2 & 28.5 \\ 
    \rowcolor{OurColor}
    \textbf{\ours (Overall Model)} & 5 & & \textbf{35.5} & &  \textbf{40.4} & \textbf{39.8} & \textbf{40.1} \\
    \hspace{0.3cm}always w/ \texttt{\RET} token & 5 & & 35.3 & & 40.2 & \textbf{39.8} & 40.0 \\
    \hspace{0.3cm}w/ Contriever passages (w/o \REL/\NOREL) & 5 & & 29.3 & & 30.8 & 29.1 & 29.9 \\
    \hspace{0.3cm}w/o \texttt{\REL/\NOREL} tokens & 5 & & 23.6 & & 32.2 & 30.6 & 31.4 \\
    \hspace{0.3cm}w/o KB (always with \NORET token) & - & & 21.3 & & 17.7 & 15.3 & 16.4 \\
    \midrule
    \rowcolor{lightgray}
    \multicolumn{8}{l}{\textit{Varying the Number of Retrieved Documents}} \\
    & 1 & & 29.0 & & \textbf{40.6} & \textbf{41.0} & \textbf{40.8} \\
    \rowcolor{OurColor}
    \textbf{\ours (Overall Model)} & 5 & & 35.5 & & 40.4 & 39.8 & 40.1 \\
     & 10 & & \textbf{36.0} & & 37.2 & 37.0 & 37.1 \\
    & 20 & & 35.7 & & 30.6 & 31.3 & 30.9 \\
  \bottomrule
  \end{tabular}
  }
  \vspace{-0.2cm}
\caption{Ablation study results demonstrating the effectiveness of the proposed reflective tokens and training strategy, along with the impact of different numbers of retrieved documents.}
  \label{tab:ablation1}
  \vspace{-0.3cm}
\end{table}

\tit{Results using Oracle Documents} To thoroughly evaluate the performance of our model, we conduct experiments under an oracle setting, where the ground-truth entity (\ie~the Wikipedia page associated with the query) is provided. In this configuration, all text passages from the oracle entity are input to \ours, which then selects the relevant passages before generating an answer. Table~\ref{tab:results_oracle} presents the results of this analysis, directly comparing \ours with Wiki-LLaVA, which leverages a Contriever model~\cite{izacard2021unsupervised} to retrieve the most relevant passages within the oracle document. We also report the performance of standard LLMs and MLLMs when prompted with the entire Wikipedia article. Notably, \ours achieves the highest performance across both the Encyclopedic-VQA and InfoSeek benchmarks, surpassing Wiki-LLaVA and standard models, further highlighting its effectiveness in isolating the passages most pertinent to the given image-question pair. It is also noteworthy that while vanilla LLMs achieve high accuracy on Encyclopedic-VQA when prompted with the entire oracle Wikipedia page, on the InfoSeek having a selection strategy of the most relevant passages lead to significantly better performance, with \ours always reaching the best results.

\begin{figure*}[t]
\begin{minipage}{0.325\linewidth}
\scriptsize{\textbf{Q}: What is one of the traditional uses of this plant?\vspace{0.05cm}}\\
\begin{minipage}{0.443\linewidth}
\includegraphics[width=1.\linewidth]{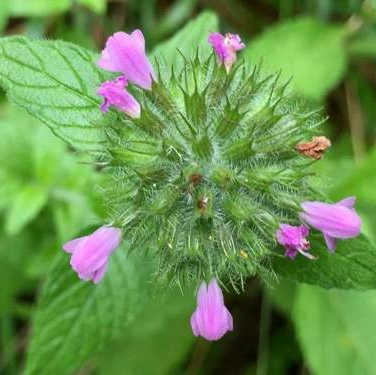}
\end{minipage}
\hfill
\begin{minipage}{0.53\linewidth}
\scriptsize{
\textbf{Wiki-LLaVA~\cite{caffagni2024wiki}}:\\
Food \textcolor{red}{\xmark} \\
\textbf{EchoSight~\cite{yan2024echosight}}:\\
Promote wound healing \textcolor{red}{\xmark} \\
\textbf{\ours (Ours):}\\
Astringent \textcolor[HTML]{00b050}{\cmark}
}
\end{minipage}
\end{minipage}
\hspace{0.02cm}
\begin{minipage}{0.325\linewidth}
\scriptsize{\textbf{Q}: Who designed this palace?\vspace{0.05cm}}\\
\begin{minipage}{0.443\linewidth}
\includegraphics[width=1.\linewidth]{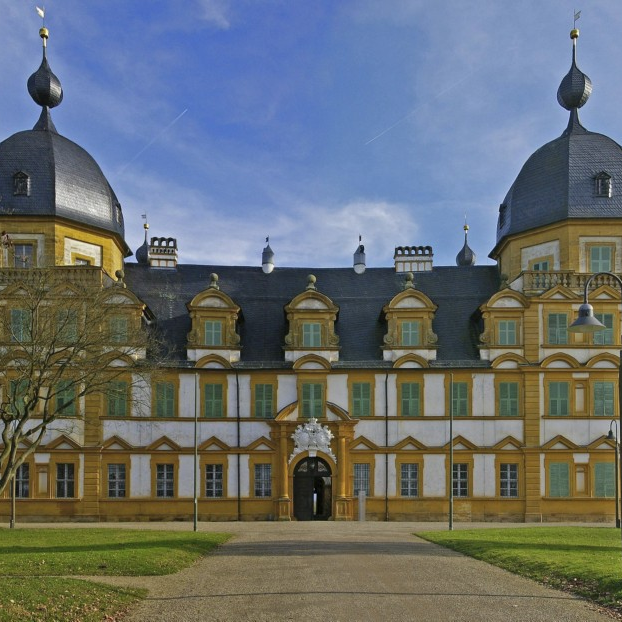}
\end{minipage}
\hfill
\begin{minipage}{0.53\linewidth}
\scriptsize{
\textbf{Wiki-LLaVA~\cite{caffagni2024wiki}}:\\
Johann Von Fischer \textcolor{red}{\xmark} \\
\textbf{EchoSight~\cite{yan2024echosight}}:\\
A team of architects, including Johan Dientzenhofer \textcolor{red}{\xmark} \\
\textbf{\ours (Ours):}\\
Balthasar Neumann \textcolor[HTML]{00b050}{\cmark}
}
\end{minipage}
\end{minipage}
\hspace{0.02cm}
\begin{minipage}{0.325\linewidth}
\scriptsize{\textbf{Q}: What was the first subspecies of this bird?\vspace{0.05cm}}\\
\begin{minipage}{0.443\linewidth}
\includegraphics[width=1.\linewidth]{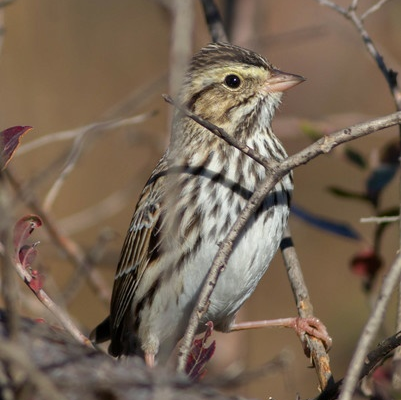}
\end{minipage}
\hfill
\begin{minipage}{0.53\linewidth}
\scriptsize{
\textbf{Wiki-LLaVA~\cite{caffagni2024wiki}}:\\
Nominate \textcolor{red}{\xmark} \\
\textbf{EchoSight~\cite{yan2024echosight}}:\\
I don't see any information about a bird species \textcolor{red}{\xmark} \\
\textbf{\ours (Ours):}\\
Aleutian Savannah Sparrow \textcolor[HTML]{00b050}{\cmark}
}
\end{minipage}
\end{minipage}
\vspace{0.1cm}

\begin{minipage}{0.325\linewidth}
\scriptsize{\textbf{Q}: What is the parent organization of this building?\vspace{0.05cm}}\\
\begin{minipage}{0.443\linewidth}
\includegraphics[width=1.\linewidth]{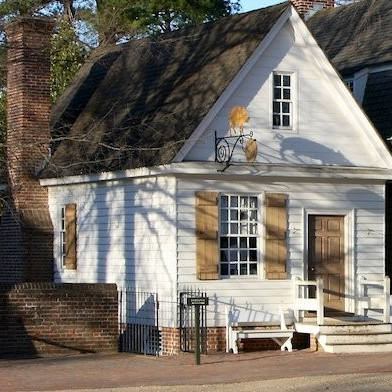}
\end{minipage}
\hfill
\begin{minipage}{0.53\linewidth}
\scriptsize{
\textbf{Wiki-LLaVA~\cite{caffagni2024wiki}}:\\
National Park Service \textcolor{red}{\xmark} \\
\textbf{EchoSight~\cite{yan2024echosight}}:\\
National Register of Historic Places \textcolor{red}{\xmark} \\
\textbf{\ours (Ours):}\\
Colonial Williamsburg Foundation \textcolor[HTML]{00b050}{\cmark}
}
\end{minipage}
\end{minipage}
\hspace{0.02cm}
\begin{minipage}{0.325\linewidth}
\scriptsize{\textbf{Q}: Which road, railway or canal does this river carry?\vspace{0.05cm}}\\
\begin{minipage}{0.443\linewidth}
\includegraphics[width=1.\linewidth]{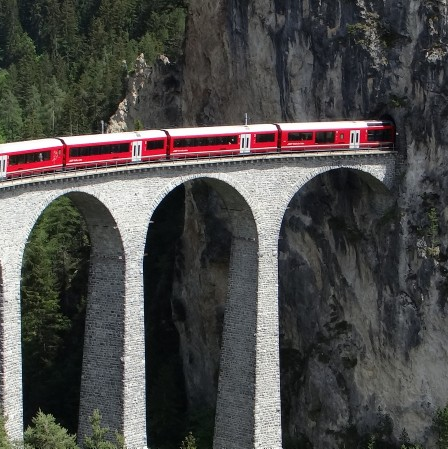}
\end{minipage}
\hfill
\begin{minipage}{0.53\linewidth}
\scriptsize{
\textbf{Wiki-LLaVA~\cite{caffagni2024wiki}}:\\
Alp Railway \textcolor{red}{\xmark} \\
\textbf{EchoSight~\cite{yan2024echosight}}:\\
Railway \textcolor{red}{\xmark} \\
\textbf{\ours (Ours):}\\
Albula Railway \textcolor[HTML]{00b050}{\cmark}
}
\end{minipage}
\end{minipage}
\hspace{0.02cm}
\begin{minipage}{0.325\linewidth}
\scriptsize{\textbf{Q:} What is the source that produces this plant?\vspace{0.05cm}}\\
\begin{minipage}{0.443\linewidth}
\includegraphics[width=1.\linewidth]{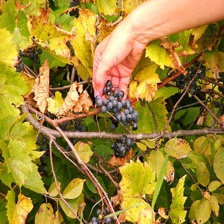}
\end{minipage}
\hfill
\begin{minipage}{0.53\linewidth}
\scriptsize{
\textbf{Wiki-LLaVA~\cite{caffagni2024wiki}}:\\
Vitis \textcolor{red}{\xmark} \\
\textbf{EchoSight~\cite{yan2024echosight}}:\\
Ephraim Wales Bull \textcolor{red}{\xmark} \\
\textbf{\ours (Ours):}\\
Vitis Lambrusca \textcolor[HTML]{00b050}{\cmark}
}
\end{minipage}
\end{minipage}
\vspace{-0.2cm}
\caption{Sample qualitative results on image-question pairs from Encyclopedic-VQA (top row) and InfoSeek (bottom row), where we compare the answers provided by \ours with those from WikiLLaVA~\cite{caffagni2024wiki} and EchoSight~\cite{yan2024echosight}.}
\label{fig:qualitatives}
\vspace{-0.14cm}
\end{figure*}

\tit{Qualitative Results}
Fig.~\ref{fig:qualitatives} provides a qualitative comparison on sample image-question pairs from Encyclopedic-VQA (top row) and InfoSeek (bottom row).

\begin{table*}[t] 
\small
\centering
\setlength{\tabcolsep}{.3em}
\resizebox{\linewidth}{!}{
\begin{tabular}{lcc ccccccccccc}
\toprule
\textbf{Model} & \textbf{LLM} & & \textbf{MMMU} & \textbf{MMB (EN)} & \textbf{POPE} & \textbf{SEED-Img} & \textbf{MME (P)} & \textbf{MME (C)} & \textbf{GQA} &  \textbf{TextVQA} & \textbf{Science-QA} & \textbf{AI2D} \\
\midrule
LLaVA-v1.5~\cite{liu2023improved} & Vicuna-7B & & 34.2 & 65.3 & 85.6 & 66.8 & 1474.3 & 314.6 & 62.4 & 58.2 & 69.0 & 56.4 \\
LLaVA-v1.5~\cite{liu2023improved} & LLaMA-3.1-8B & & 39.4 & 72.4 & 85.1 & 69.8 & 1531.5 & 353.3 & 63.6 & 58.4 & 76.3 & 61.8 \\
\midrule
Wiki-LLaVA (E-VQA)~\cite{caffagni2024wiki} & Vicuna-7B & & 36.6 & 70.4 & 86.6 & - & 1170.1 & 290.0 & - & - & - & - \\
Wiki-LLaVA (InfoSeek)~\cite{caffagni2024wiki} & Vicuna-7B & & 35.6 & 71.1 & 84.2 & - & 1438.9 & 341.3 & - & - & - & - \\
Wiki-LLaVA (E-VQA)~\cite{caffagni2024wiki}$^\diamondsuit$ & LLaMA-3.1-8B & & 32.2 & 60.9 & 84.6 & 59.2 & 1350.7 & 306.8 & 56.6 & 49.1 & 67.5 & 55.1 \\
Wiki-LLaVA (InfoSeek)~\cite{caffagni2024wiki}$^\diamondsuit$ & LLaMA-3.1-8B & & 35.9 & 52.0 & 85.7 & 60.5 & 1417.8 & 349.6 & 58.6 & 50.1 & 69.1 & 54.3 \\
\rowcolor{OurColor}
\textbf{\ours (Ours)} & LLaMA-3.1-8B & & 38.9 & 69.9 & 85.1 & 68.6 & 1564.5 & 355.7 & 62.1 & 56.8 & 75.4 & 60.6 \\
\bottomrule
\end{tabular}
}
\vspace{-0.2cm}
\caption{Performance preservation analysis on standard benchmarks for MLLM evaluation and traditional VQA datasets.}
\vspace{-0.35cm}
\label{tab:mllms}
\end{table*}

\subsection{Ablation Studies and Analyses}

\tinytit{Effectiveness of Two-Stage Training} We analyze the impact of our two-stage, two-model training strategy. Specifically, we assess the effectiveness of the in-article reflective model, which is trained using both positive and negative passages drawn from the same Wikipedia pages, on Encyclopedic-VQA and InfoSeek. Table~\ref{tab:ablation1} (top) demonstrates that training only the first model alone does not yield high accuracy scores. Likewise, training a single model on full-dataset annotations generated by open-source LLMs such as LLaMA-3.1 fails to achieve competitive performance. In constrast, the complete model (\ie~\ours) consistently achieves higher accuracy.

\tit{Effectiveness of Reflective Tokens} To evaluate the effectiveness of the proposed reflective tokens, we conduct ablation studies that isolate the contributions of both retrieval and relevance tokens. We design four variants of the inference pipeline: (i) always enforcing the \RET token while using the standard \REL/\NOREL pipeline, (ii) without using relevance tokens, instead selecting the top passages retrieved by the Contriever~\cite{izacard2021unsupervised} model, (iii) without assessing the relevance of retrieved passages, where two random passages are selected from each of the top-5 documents, and (iv) without retrieval from the knowledge base (\ie~always enforcing \NORET during generation). Results are summarized in the top part of Table~\ref{tab:ablation1} for Encyclopedic-VQA and InfoSeek. As shown, consistently performing retrieval causes only a minor performance degradation, given that both datasets are designed to require external knowledge. In contrast, omitting the proposed relevance tokens leads to a substantial accuracy drop (\eg~from 35.5 to 23.6 on Encyclopedic-VQA and from 40.1 to 31.4 on InfoSeek), underscoring the critical role of identifying relevant passages before answer generation. Furthermore, the lowest scores occur when retrieval from the external knowledge base is entirely bypassed, emphasizing the need for retrieval. Also, replacing the \REL and \NOREL tokens with the top-2 passages retrieved by Contriever results in significantly lower performance on both datasets, underscoring the benefit of explicit relevance modeling over direct retrieval.

\tit{Varying the Number of Retrieved Documents} In Table~\ref{tab:ablation1} (bottom), we further analyze the effect of varying the number $k$ of retrieved documents. While setting $k$ equal to 10 achieves optimal results on Encyclopedic-VQA, using only the top-1 retrieved document generally performs better for InfoSeek. This discrepancy can be attributed to differences in the knowledge base sizes (\ie~significantly larger for Encyclopedic-VQA) and retrieval performance across the two datasets (cf. Table~\ref{tab:retrieval_small}). Overall, using the top-5 retrieved documents provides the best trade-off across both datasets, leading us to set $k$ equal to 5 in our experiments.

\tit{Performance Preservation on Standard Benchmarks} 
Finally, we evaluate whether the proposed approach impacts performance on standard MLLM and VQA benchmarks that do not require external knowledge. In Table~\ref{tab:mllms}, we compare \ours with Wiki-LLaVA, which fine-tunes a LLaVA model for similar purposes, and include results from the original LLaVA-v1.5 model, tested with both Vicuna-7B and LLaMA-3.1. \ours incurs only a minor performance reduction relative to the original LLaVA model, while significantly outperforming Wiki-LLaVA.

\section{Conclusion}
\label{sec:conclusion}
We proposed \ours, a multimodal LLM with retrieval-augmented generation. Our method employs reflective tokens, trained in a two-stage two-model pipeline. Extensive experiments, conducted on both VQA datasets requiring external knowledge and standard datasets, demonstrate the efficacy of the proposed solution.

\section*{Acknowledgments}
We acknowledge the CINECA award under the ISCRA initiative, for the availability of high-performance computing resources and support. This work has been conducted under a research grant co-funded by Altilia s.r.l., and supported by the PNRRM4C2 project ``FAIR - Future Artificial Intelligence Research'', funded by the European Commission, by the PNRR project ``Italian Strengthening of Esfri RI Resilience'' (ITSERR), funded by the European Union - NextGenerationEU (CUP B53C22001770006), and by the ``Fortissimo Plus'' (FFplus) project, funded by the European High-Performance Computing Joint Undertaking (No. 101163317). We would also like to thank Davide Caffagni and Sara Sarto for their valuable support.

{
    \small
    \bibliographystyle{ieeenat_fullname}
    \bibliography{bibliography}
}

\clearpage
\maketitlesupplementary
\appendix
\begin{table*}[bp]
    \vspace{-0.2cm}
    \centering
    \setlength{\tabcolsep}{.4em}
    \resizebox{\linewidth}{!}{
    \begin{tabular}{lc cccccc}
    \toprule
     & & \textbf{Wiki-LLaVA~\cite{caffagni2024wiki}} &\textbf{RORA-VLM~\cite{qi2024rora}} & \textbf{EchoSight~\cite{yan2024echosight}} & \textbf{Wiki-LLaVA~\cite{caffagni2024wiki}$^\diamondsuit$} & \textbf{EchoSight~\cite{yan2024echosight}$^\diamondsuit$} & \cellcolor{OurColor}\textbf{\ours (Ours)} \\
     \midrule
     \rowcolor{lightgray} 
     \multicolumn{8}{l}{\textit{Architectural Design}} \\
     LLM & & Vicuna-7B & Vicuna-7B & Mistral-7B/LLaMA-3-8B & LLaMA-3.1-8B & LLaMA-3.1-8B & \cellcolor{OurColor}LLaMA-3.1-8B \\
     Underlying MLLM & & LLaVA-v1.5 & LLaVA-v1.5 & - & LLaVA-v1.5 & - & \cellcolor{OurColor}LLaVA-v1.5 \\
     LLM Fine-tuning & & \cmark & \cmark & \xmark & \cmark & \xmark & \cellcolor{OurColor}\cmark \\
     Inherently Multimodal & & \cmark & \cmark & \xmark & \cmark & \xmark & \cellcolor{OurColor}\cmark \\
     Dataset Independent & & \xmark & \xmark & \xmark & \xmark & \cmark & \cellcolor{OurColor}\cmark \\
     \midrule
     \rowcolor{lightgray} 
     \multicolumn{8}{l}{\textit{Encyclopedic-VQA}} \\
     \# KB Items & & 2M & N/A & 2M & 2M & 2M & \cellcolor{OurColor}2M \\
     From the Original KB? & & \cmark & \xmark~(WIT KB) & \cmark & \cmark & \cmark & \cellcolor{OurColor}\cmark \\
     \midrule
     \rowcolor{lightgray} 
     \multicolumn{8}{l}{\textit{InfoSeek}} \\
     \# KB Items & & 100k & N/A & 100k & 100k & 100k & \cellcolor{OurColor}100k \\
     From the Original KB? & & \cmark & \xmark~(WIT KB) & \xmark~(E-VQA KB) & \cmark & \cmark & \cellcolor{OurColor}\cmark \\
    \bottomrule
    \end{tabular}
    }
    \vspace{-0.2cm}
    \caption{Comparison with existing methods based on LLMs in terms of architectural design and knowledge bases used for the Encyclopedic-VQA and InfoSeek datasets. The marker $\diamondsuit$ represents our reproductions.}
    \label{tab:comparison}
\end{table*}

\noindent In the following, we provide additional details and analyses for the proposed Reflective LLaVA (\ours) model. First, we outline key distinctions between \ours and existing approaches, focusing on architectural design and the knowledge bases used. Next, we present a comprehensive overview of the implementation details, training setup, and the data mixture employed in our two-stage, two-model training strategy. Finally, we supplement the main paper with extended experimental evaluations and qualitative results, further validating the effectiveness of our approach.

\section{Detailed Comparison with Existing Methods}
\tinytit{Positioning} Knowledge-based VQA has been widely studied by the Computer Vision community, particularly with the emergence of retrieval-augmented architectures~\cite{lin2022revive,gui2022kat,hu2023reveal,lin2023fine} tailored for small-scale datasets such as OK-VQA~\cite{marino2019ok,schwenk2022okvqa}. More recently, the introduction of larger-scale and challenging datasets like Encyclopedic-VQA~\cite{mensink2023encyclopedic} and InfoSeek~\cite{chen2023can}, alongside advances in LLMs, has shifted the focus towards developing retrieval-augmented solutions leveraging LLMs and MLLMs. In this context, addressing knowledge-based VQA at a Wikipedia-scale remains a relatively unexplored challenge. To the best of our knowledge, only a few methods have attempted to tackle this task effectively. Among these, Wiki-LLaVA~\cite{caffagni2024wiki}, RORA-VLM~\cite{qi2024rora}, and EchoSight~\cite{yan2024echosight} are the most comparable to our approach. Table~\ref{tab:comparison} presents a detailed comparison of \ours with these state-of-the-art LLM-based methods specifically designed for this task.

\tit{Architectural Design} As detailed in the main paper, both RORA-VLM and Wiki-LLaVA are built upon the LLaVA-v1.5 MLLM, using Vicuna-7B as LLM. In contrast, EchoSight employs different LLMs depending on the target dataset (\ie~Mistral-7B for Encyclopedic-VQA and LLaMA-3-8B for InfoSeek). Our proposed \ours model, instead, is based on the LLaVA-v1.5 model with LLaMA-3.1-8B as the underlying LLM. To ensure a fair comparison and eliminate any advantage from using a more advanced LLM, we additionally implement reproductions of Wiki-LLaVA and EchoSight using LLaMA-3.1-8B. 

Regarding the training phase, EchoSight integrates a re-ranking component to reorder retrieved passages, which is specifically trained on Encyclopedic-VQA while keeping the LLM frozen. In contrast, the other competitors are fine-tuned on the considered datasets. Furthermore, unlike RORA-VLM and Wiki-LLaVA, which train separate models for Encyclopedic-VQA and InfoSeek, \ours is fine-tuned jointly on both datasets, enabling seamless applicability across both settings.

\tit{Knowledge Bases} Each considered dataset is paired with its own knowledge base. Specifically, Encyclopedic-VQA is associated with a controlled knowledge base comprising 2 million English articles, derived from the WIT dataset~\cite{srinivasan2021wit}. Similarly, InfoSeek utilizes a knowledge base of Wikipedia pages, initially consisting of 100k items (as reported in the original paper~\cite{chen2023can}) and later expanded to 6 million entities\footnote{\scriptsize\url{https://github.com/open-vision-language/infoseek}}. With the exception of RORA-VLM, which uses its own knowledge base directly extracted from WIT, all other approaches adhere to the original evaluation protocols. Specifically, experiments on Encyclopedic-VQA are conducted with the original knowledge base of 2 million entities. For InfoSeek, following~\cite{chen2023can}, evaluations are conducted using a subset of 100k Wikipedia pages. While both Wiki-LLaVA and \ours extract subsets directly from the 6M knowledge base provided with the InfoSeek dataset, EchoSight uses 100k entities extracted from the Encyclopedic-VQA knowledge base. To ensure a fair comparison, our reproductions of Wiki-LLaVA and EchoSight are tested using the same knowledge bases employed in our approach.

\begin{table}[t]
    \centering
    \setlength{\tabcolsep}{.2em}
    \resizebox{\linewidth}{!}{
    \begin{tabular}{lc cc c cc}
    \toprule
        & & \multicolumn{2}{c}{\textbf{First Stage}} & & \multicolumn{2}{c}{\textbf{Second Stage}} \\
        \cmidrule{3-4} \cmidrule{6-7}
        & & \# Samples & Passages & & \# Samples & Passages \\
        \midrule
         E-VQA & & 43.6k & In-Article & & 2.9M & In- and Cross-Article \\
        InfoSeek & & 41.0k & In-Article & & 2.5M & In- and Cross-Article \\
        LLaVA-Instruct & &  665.3k & - & &  665.3k & - \\
    \bottomrule
    \end{tabular}
    }
    \vspace{-0.2cm}
    \caption{Training data mixture employed during the two phases of the proposed training strategy.}
    \label{tab:data_mixture}
    \vspace{-0.3cm}
\end{table}

\section{Additional Experimental Details}

\subsection{Datasets}

\tinytit{Training Data Mixture} In Table~\ref{tab:data_mixture}, we summarize the training datasets employed in both stages of the proposed training strategy. At each stage, the samples from each data source are balanced to maintain consistent representativeness. The annotations for relevant and non-relevant passages will be made publicly available.

\tit{Additional Knowledge-based VQA Datasets} In addition to the datasets employed in the main paper, we perform zero-shot experiments on two additional knowledge-based VQA datasets, namely ViQuAE~\cite{lerner2022viquae} and S3VQA~\cite{jain2021select}. Specifically, we report experiments on the ViQuAE test set, composed of 3,317 image-questions pairs, and on the S3VQA validation set, that contains 750 samples. To facilitate the experiments, for both datasets, we employ the 2M knowledge base of Encyclopedic-VQA, performing image-to-image retrieval with EVA-CLIP as the retrieval model. To compute the results, for ViQuAE samples we follow the official evaluation protocol which includes F1 and exact match scores. Instead, given the high level of difficulty of the questions contained in S3VQA and the absence of an official evaluation protocol, we evaluate generated answers using GPT-4. In detail, following recent literature~\cite{chan2023clair,wu2024clair,moratelli2024revisiting}, we prompt GPT-4 with the question and the image description extracted with BLIP-2~\cite{li2023blip}\footnote{\scriptsize\url{https://huggingface.co/Salesforce/blip2-flan-t5-xl}}, and ask the model to evaluate the alignment between the predicted and ground-truth answers with a score from 0 to 100, where 0 indicates no alignment and 100 indicates a perfect alignment between the two answers\footnote{ 
\footnotesize{Specifically, the prompt we used in our evaluation is:}

\noindent\scriptsize{\texttt{You are trying to evaluate the alignment between a predicted answer and a ground-truth answer for a given question-image pair. To do this, consider the context provided by the question itself and the caption of the query image.}}

\noindent\scriptsize{\texttt{\#~Question:~\{question\}}}

\noindent\scriptsize{\texttt{\#~Image Caption:~\{caption\}}}

\noindent\scriptsize{\texttt{\#~Ground-truth Answer:~\{ground-truth answer\}}}

\noindent\scriptsize{\texttt{\#~Predicted Answer:~\{predicted answer\}}}

\noindent\scriptsize{\texttt{You have to determine the alignment between the predicted answer and the ground-truth on a scale from 0 to 100, where 0 indicates no alignment and 100 indicates perfect alignment. Your response should be in JSON format, outputting a list where each element is a dictionary representing a candidate with:}}

\noindent\scriptsize{\texttt{``score'': a numeric value between 0 and 100 indicating the alignment level,}}

\noindent\scriptsize{\texttt{``reason'': a string explaining the rationale for the given score.}}
}.

\tit{Benchmarks for MLLM Evaluation} Table~\ref{tab:mllms} of the main paper presents the performance of \ours on a suite of standard benchmarks commonly used to evaluate MLLMs. The evaluation includes results on MMMU~\cite{yue2023mmmu}, the English subset of MMBench (MMB)~\cite{liu2023mmbench}, POPE~\cite{li2023evaluating}, the image-specific subset of SEED-Bench (SEED-Img)~\cite{li2023seed}, and MME~\cite{fu2023mme}. These benchmarks comprehensively assess MLLM capabilities across task-oriented and instruction-following scenarios. Additionally, we report results on established VQA datasets, including GQA~\cite{hudson2019gqa}, TextVQA~\cite{singh2019towards}, ScienceQA~\cite{lu2022learn}, and AI2D~\cite{kembhavi2016diagram}. All evaluations were performed using a dedicated library tailored for benchmarking large multimodal models\footnote{\scriptsize\url{https://github.com/EvolvingLMMs-Lab/lmms-eval}}.

\subsection{Additional Implementation Details}
In our experiments, we adopt the prompt formats used by LLaMA-3.1 and utilize three of the four supported roles: \texttt{\small{system}}, \texttt{\small{user}}, and \texttt{\small{assistant}}. The instructions provided are consistent with the style of the LLaVA-v1.5 MLLM. Specifically, for each image-question pair, the prompt we employ is as follows:
\begin{lstlisting}
(*\ttfamily\bfseries\color{GrayL}<|begin\_of\_text|>*)
(*\bfseries\ttfamily\color{GrayL} <|start\_header\_id|>system<|end\_header\_id|>*)
    You are a helpful language and vision assistant. You are able to understand the visual content that the user provides, and assist the user with a variety of tasks 
    using natural language.(*\ttfamily\bfseries\color{GrayL}<|eot\_id|>*)
    
(*\bfseries\ttfamily\color{cornellred}<|start\_header\_id|>user<|end\_header\_id|>*)
    (*\bfseries\ttfamily\color{imagecolor}<image>*)
    What color is the car?(*\ttfamily\bfseries\color{GrayL}<|eot\_id|>*)

(*\bfseries\ttfamily\color{coolblack}<|start\_header\_id|>assistant<|end\_header\_id|>*)
(*\color{GrayL}-------------------------------------------------*)
    (*\bfseries\ttfamily\color{noretcolor}<NORET>*)
    Black(*\ttfamily\bfseries\color{GrayL}<|eot\_id|>*)
\end{lstlisting}
In practice, \texttt{\small{<image>}} is replaced with the visual tokens generated by the visual encoder, after being projected into the space of the LLM using the vision-to-language adapter of the model (\ie~an MLP). For completeness, we report the output of the model below the dashed line. In this case, the query does not require retrieval; therefore, \ours generates the \NORET token and directly provides the answer.

When instead the image-question pair requires retrieval to effectively produce the final answer, the model first generates the \RET token. After performing retrieval and allowing the model to identify the relevant passages, the answer is finally generated. Specifically, the complete prompt after the retrieval stage is as follows: 

\begin{lstlisting}
(*\ttfamily\bfseries\color{GrayL}<|begin\_of\_text|>*)
(*\bfseries\ttfamily\color{GrayL} <|start\_header\_id|>system<|end\_header\_id|>*)
    You are a helpful language and vision assistant. You are able to understand the visual content that the user provides, and assist the user with a variety of tasks 
    using natural language.(*\ttfamily\bfseries\color{GrayL}<|eot\_id|>*)
    
(*\bfseries\ttfamily\color{cornellred}<|start\_header\_id|>user<|end\_header\_id|>*)
    (*\bfseries\ttfamily\color{imagecolor}<image>*)
    How big can this plant become?(*\ttfamily\bfseries\color{GrayL}<|eot\_id|>*)

(*\bfseries\ttfamily\color{coolblack}<|start\_header\_id|>assistant<|end\_header\_id|>*)
    (*\bfseries\ttfamily\color{retcolor}<RET>*)

(*\bfseries\ttfamily\color{cornellred}<|start\_header\_id|>user<|end\_header\_id|>*)
    Consider this paragraph:
    (*\bfseries\ttfamily\color{passagecolor}<paragraph>*)
    Prunus laurocerasus is an evergreen shrub 
    or small to medium-sized tree, growing to
    5 to 15 metres (16 to 49ft) tall, rarely to 18 metres (59ft), with a trunk up to 60cm broad. The leaves are dark green, leathery, shiny, with a finely serrated margin. The leaves can have the scent of almonds when crushed. The flower buds appear in early spring and open in early summer in erect
    7 to 15cm racemes of 40 flowers, each 
    flower 1cm across, with five creamy-white petals and numerous yellowish stamens with
    a sweet smell.
    (*\bfseries\ttfamily\color{passagecolor}</paragraph>*)
    Give a short answer.(*\ttfamily\bfseries\color{GrayL}<|eot\_id|>*)

(*\bfseries\ttfamily\color{coolblack}<|start\_header\_id|>assistant<|end\_header\_id|>*)
(*\color{GrayL}-------------------------------------------------*)
    (*\bfseries\ttfamily\color{relcolor} <REL>*)
    16 to 49ft(*\ttfamily\bfseries\color{GrayL}<|eot\_id|>*)
\end{lstlisting}
For simplicity, here we include a single textual passage. However, as detailed in Sec.~\ref{sec:tokens} of the main paper, the final answer is generated using the set of passages that have been judged relevant. As shown, we also introduce two additional special tokens, \texttt{\small{<paragraph>}} and \texttt{\small{</paragraph>}}, to mark the beginning and end of each passage. These tokens are incorporated into the original LLaMA-3.1 vocabulary, as described for the proposed reflective tokens (Sec.~\ref{sec:setup}).

\subsection{Baselines and Competitor Reproductions}
\tinytit{Vanilla LLMs and MLLMs} To evaluate text-only and multimodal LLMs, we adopt the original system prompts provided by the models, appending the instruction ``\texttt{\small{Give a short answer}}'' to enforce concise responses. For text-only LLMs, we supplement the input with automatically generated image descriptions extracted using BLIP-2~\cite{li2023blip}.

\tit{Wiki-LLaVA~\cite{caffagni2024wiki}} Following the original paper, all experiments involving the Wiki-LLaVA model are conducted with the number $k$ of retrieved documents equal to 1, where the LLM is provided with the top-3 passages retrieved via the Contriever model~\cite{izacard2021unsupervised}. To evaluate Wiki-LLaVA with the LLaMA-3.1 LLM, we train two separate models as described in the original work: one on the Encyclopedic-VQA training set and the other on the InfoSeek training data, adhering to the same experimental settings.

\tit{EchoSight~\cite{yan2024echosight}} Experiments with LLaMA-3.1 are conducted by replacing the original LLM (\ie~Mistral for Encyclopedic-VQA and LLaMA-3 for InfoSeek) while keeping all other experimental settings unchanged. The LLM is prompted as outlined in the original paper, employing a one-shot example for InfoSeek experiments.

\begin{figure}[t]
    \centering
    \includegraphics[width=0.99\linewidth]{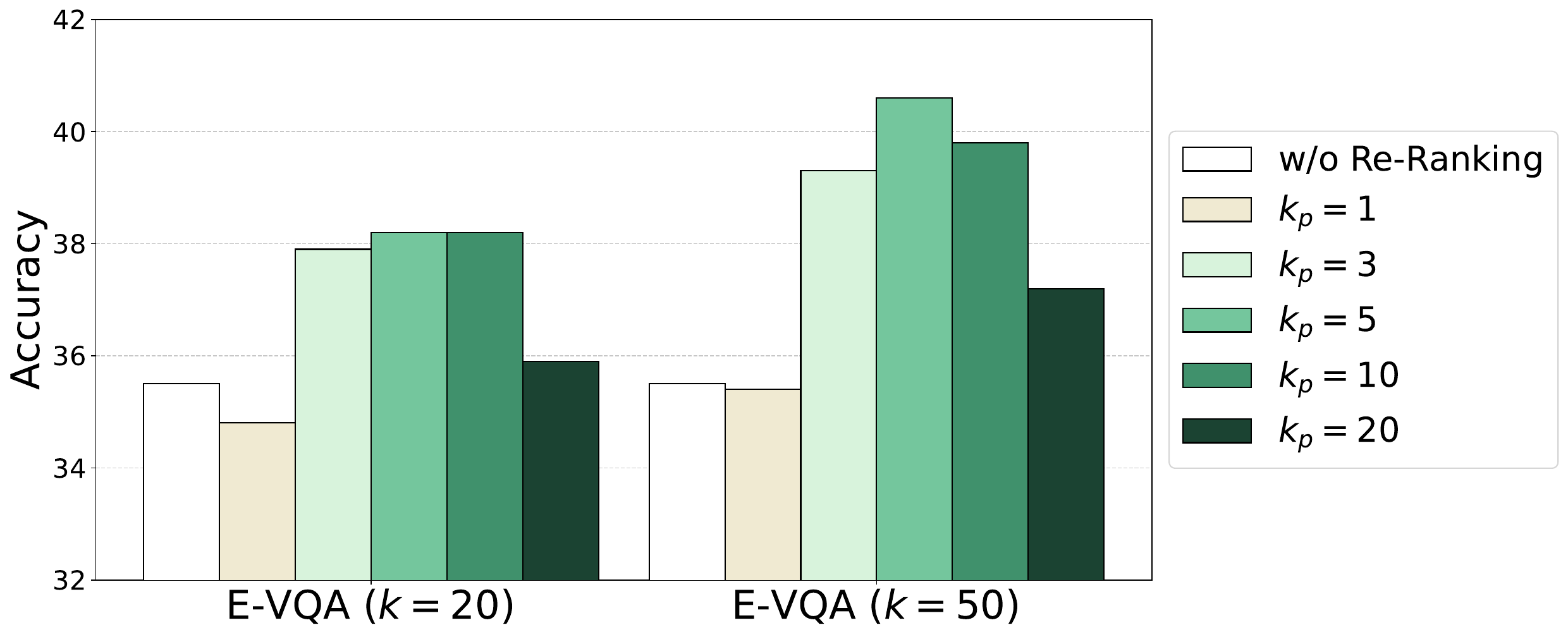}
    \vspace{-.2cm}
    \caption{Re-ranking performance on the single-hop question split of Encyclopedic-VQA when varying the number $k$ of retrieved documents and the number $k_p$ of passages given to our model after re-ranking. The ``without re-ranking'' bars refer to the best results achieved without using the re-ranking component, as reported in Table~\ref{tab:results} of the main paper.}
    \label{fig:reranking}
    \vspace{-.15cm}
\end{figure}

\begin{table}[t]
  \centering
  \setlength{\tabcolsep}{.2em}
  \resizebox{\linewidth}{!}{
  \begin{tabular}{lcc cc c cc}
   \toprule
    & & & \multicolumn{2}{c}{\textbf{ViQuAE}} & & \multicolumn{1}{c}{\textbf{S3VQA}} \\
    \cmidrule{4-5} \cmidrule{7-7}
    \textbf{Model} & \textbf{LLM} & & F1 & EM & & GPT-4 \\
    \midrule
    LLaVA-v1.5~\cite{liu2023improved} & Vicuna-7B & & 15.1 & 26.6 & & 23.9 \\
    LLaVA-v1.5~\cite{liu2023improved} & LLaMA-3.1-8B & & 15.0 & 25.6 & & 24.4 \\
    \midrule
     Wiki-LLaVA (E-VQA)~\cite{caffagni2024wiki}$^\diamondsuit$ & LLaMA-3.1-8B & & 10.5 & 16.7 & & 22.7 \\
    Wiki-LLaVA (InfoSeek)~\cite{caffagni2024wiki}$^\diamondsuit$ & LLaMA-3.1-8B  & & 12.7 & 21.8 & & 21.8 \\
    \textbf{\ours} (w/o KB) & LLaMA-3.1-8B & & 16.6 & 27.6 & & 26.9 \\
    \rowcolor{OurColor}
     &  & & \textbf{23.2} & \textbf{38.1} & & \multicolumn{1}{c}{\textbf{29.3}} \\
    \rowcolor{OurColor} 
    \multirow{-2}{*}{\textbf{\ours (Ours)}} & \multirow{-2}{*}{LLaMA-3.1-8B}  & & \multicolumn{2}{c}{\textcolor[HTML]{00b050}{(52.0\%)}} & & \textcolor[HTML]{00b050}{(16.8\%)} \\
  \bottomrule
  \end{tabular}
  }
  \vspace{-0.2cm}
\caption{Zero-shot performance on additional knowledge-based VQA datasets. The percentage of samples in which our model incorporates external knowledge is highlighted in \textcolor[HTML]{00b050}{green}.}
\vspace{-0.4cm}
  \label{tab:kvqa_datasets}
\end{table}

\section{Additional Experimental Results} 

\subsection{Further Ablation Studies and Analyses}

\tinytit{Results on Other Knowledge-based Datasets}
In addition to the results on the Encyclopedic-VQA and InfoSeek datasets, we also validate the generalization capabilities of \ours to zero-shot settings which always require knowledge retrieval. Specifically, we report the results on two additional knowledge-based VQA datasets, \ie~ViQuAE~\cite{lerner2022viquae} and S3VQA~\cite{jain2021select}. 
From Table~\ref{tab:kvqa_datasets}, it can be seen that even in these challenging settings, \ours achieves the best results, outperforming competitors by a significant margin and demonstrating the usefulness of predicting reflective tokens. 

\tit{Re-Ranking Analysis when Varying $k$ and $k_p$}
As a complement to the experiments with the re-ranking component shown in Table~\ref{tab:results_reranking}, we report in Fig.~\ref{fig:reranking} the performance of \ours as a function of the number $k$ of retrieved documents and the number $k_p$ of passages provided to the model after re-ranking. For this experiment, we employ the re-ranker module proposed in~\cite{yan2024echosight}, trained on the Encyclopedic-VQA dataset\footnote{As mentioned in the main paper and shown in Table~\ref{tab:comparison}, re-ranking results can not be reported for the InfoSeek dataset, as the original knowledge base used in our experiments differs significantly from the one employed in~\cite{yan2024echosight}, which was derived from Encyclopedic-VQA data.}. Specifically, we use $k=\{20, 50\}$ and $k_p=\{1, 3, 5, 10, 20\}$, while also reporting the performance of our best configuration without re-ranking. All experiments are conducted using EVA-CLIP as the retrieval model with image-to-image similarity, as it demonstrates superior performance in the Encyclopedic-VQA setting (cf. Table~\ref{tab:retrieval_small}). As expected, increasing the number $k$ of retrieved documents generally improves performance, demonstrating that incorporating a re-ranking stage effectively enhances the results. However, it is worth noting that relying solely on the top-1 retrieved passage does not yield the best overall performance. Instead, the highest accuracy scores are achieved with $k_p=5$\footnote{All experiments reported in Table~\ref{tab:results_reranking} of the main paper are conducted using $k=50$ and $k_p=5$.}. These results highlight the ability of \ours to accurately identify the most relevant passages and effectively utilize them to provide more accurate answers. 

\begin{table}[t]
  \centering
  \setlength{\tabcolsep}{.35em}
  \resizebox{\linewidth}{!}{
  \begin{tabular}{lc cc c ccc}
   \toprule
    & & \multicolumn{2}{c}{\textbf{E-VQA}} & & \multicolumn{3}{c}{\textbf{InfoSeek}} \\
    \cmidrule{3-4} \cmidrule{6-8}
     \textbf{Model} & & Single-Hop & All & & Unseen-Q & Unseen-E & All \\
    \midrule
    \rowcolor{lightgray} 
    \multicolumn{8}{l}{\textit{Textual Retrieval Mode}} \\
     Self-RAG~\cite{asaiself} & & 17.9 & 17.6 & & 5.0 & 5.5 & 4.5 \\
      \rowcolor{OurColor}
    \textbf{\ours (Ours)} & & \textbf{28.0} & \textbf{29.2} & & \textbf{40.4} & \textbf{39.8} & \textbf{40.1} \\
    \midrule
    \rowcolor{lightgray} 
    \multicolumn{8}{l}{\textit{Visual Retrieval Mode}} \\
    Self-RAG~\cite{asaiself} & & 18.8 & 18.4 & & 5.1 & 4.3 & 4.6 \\
      \rowcolor{OurColor}
    \textbf{\ours (Ours)} & & \textbf{35.5} & \textbf{35.5} & & \textbf{28.6} & \textbf{28.1} & \textbf{28.3} \\
  \bottomrule
  \end{tabular}
  }
  \vspace{-0.2cm}
\caption{Experimental comparison with Self-RAG on the Encyclopedic-VQA test set and the InfoSeek validation set.}
\label{tab:results_selfrag}
\vspace{-0.3cm}
\end{table}

\tit{Comparison with Self-RAG~\cite{asaiself}} 
In addition to task-specific competitors, we also compare our model with Self-RAG~\cite{asaiself} that incorporates special tokens to improve retrieval-augmented generation pipelines. In particular, this model has been designed for natural language understanding tasks, like open-ended question answering, reasoning, and fact verification, and has no multimodal capabilities. To conduct the comparison, we adapt Self-RAG to our setting by employing the same retrieval pipeline as our model. After retrieving the top-$k$ documents\footnote{For consistency with our setting without re-ranking, we use $k=5$.}, we prompt the LLM with all passages from the retrieved documents, allowing the model to identify the most useful information for answering the input question. Since the underlying LLM is not equipped to process visual inputs, we include the image description extracted using BLIP-2~\cite{li2023blip} in the input prompt, as done for vanilla LLMs. Results on both Encyclopedic-VQA and InfoSeek are presented in Table~\ref{tab:results_selfrag}. Notably, \ours consistently outperforms Self-RAG across all settings, further highlighting the effectiveness of our approach and the critical role of incorporating multimodal information for solving the task.

\begin{table}[t] 
\small
\centering
\setlength{\tabcolsep}{.18em}
\resizebox{\linewidth}{!}{
\begin{tabular}{lc cc c c c c c cc}
\toprule
& & \multicolumn{2}{c}{\RET} & & \multicolumn{1}{c}{\NORET} & & \multicolumn{1}{c}{\REL}  & & \multicolumn{2}{c}{\NOREL} \\
\cmidrule{3-4} \cmidrule{6-6} \cmidrule{8-8} \cmidrule{10-11}
& & \multirow{2}{*}{\textbf{E-VQA}} & \multirow{2}{*}{\textbf{InfoSeek}} & & \multirow{2}{*}{\textbf{GQA}} & & \textbf{E-VQA} & & \textbf{E-VQA} & \textbf{E-VQA} \\
& &  &  & &  & & \textbf{(Pos)} & & \textbf{(Soft)} & \textbf{(Hard)} \\
\midrule
GPT-4~\cite{achiam2023gpt} & & 82.5 & 73.5 & & 94.5 & & 93.8 & & 93.4 & 91.3 \\
GPT-4V~\cite{achiam2023gpt} & & \textbf{94.4} & 96.3 & & 96.6 & & 94.4 & & 92.4 & 94.3 \\
\midrule
After LLaVA 1st stage & & 80.6 & 99.7 & & \textbf{100.0} & & 93.4 & & \textbf{96.8} & 94.8 \\
After LLaVA 2nd stage & & 88.4 & \textbf{100.0} & & \textbf{100.0} & & \textbf{94.6} & & 95.9 & \textbf{96.2} \\
\bottomrule
\end{tabular}
}
\vspace{-0.2cm}
\caption{Accuracy scores achieved by \ours in predicting the proposed reflective tokens, compared with the performance of GPT-4 and GPT-4V.}
\vspace{-0.3cm}
\label{tab:tokens}
\end{table}

\tit{Evaluating Reflective Token Accuracy} To better analyze the performance of our model, we assess its ability to correctly predict the introduced reflective tokens. Specifically, we evaluate the accuracy of the \RET and \NORET tokens by extracting a subset of 11k image-question pairs from the validation sets of Encyclopedic-VQA, InfoSeek, and GQA. For this experiment, we assume that all samples from Encyclopedic-VQA and InfoSeek necessitate retrieval, while all image-question pairs in GQA, which contain generic questions about the image content, can be answered without external retrieval. Additionally, we evaluate the prediction accuracy of the \REL and \NOREL tokens on a subset of the Encyclopedic-VQA validation set consisting of 500 image-question pairs, where the relevant textual snippet containing the answer is available. For each sample in this subset, we extract the relevant passage containing the snippet with the answer, along with two non-relevant passages. Specifically, we extract a soft negative passage from an unrelated document and a hard negative passage from the same document, ensuring that the relevant textual snippet is not included in the hard negative. 

Accuracy scores are shown in Table~\ref{tab:tokens}, where we compare the performance of zero-shot models, such as GPT-4 and GPT-4V, with the results obtained by applying our strategy after either the first or second stage of LLaVA training. LLaVA-v1.5 follows a two-stage training process: the first stage pre-trains on image-caption pairs to align image features with the LLM textual space, while the second stage focuses on enhancing multimodal conversational capabilities. We therefore analyze the optimal point in this pipeline to incorporate our strategy for learning new special tokens. As it can be seen, accuracy scores are consistently higher than 85\% for all reflective tokens, highlighting the ability of the model to accurately predict when retrieval is necessary and whether the retrieved passages are relevant to the query. Also, applying our training strategy after the second stage of LLaVA training generally yields the best results.

\subsection{Does \ours Integrate Built-In Re-Ranking Capabilities?}
As a final analysis, we evaluate whether our model can be directly employed to re-rank retrieved passages. To this end, we first retrieve the top-$k$ most relevant documents from the external knowledge base and use the log probabilities of \REL and \NOREL tokens to rank the associated textual passages. Specifically, for each passage, we compute the difference between the log probability of the \REL token and that of the \NOREL token, ranking all passages in descending order based on the computed difference scores. We then provide the top-$k_p$ passages as input to the model to generate the final answer. This analysis is conducted on the Encyclopedic-VQA dataset, with results reported in Table~\ref{tab:reranking_probs}. As it can be seen, increasing the number $k$ of retrieved documents and directly leveraging the probabilities of \REL and \NOREL tokens to rank passages leads to the best results. Notably, \ours with the built-in re-ranking strategy achieves 37.8 accuracy points on single-hop questions using $k$ equal to 50 and $k_p$ equal to 2, outperforming the best result without re-ranking by 2.3 points. While training a dedicated re-ranking module could further improve performance, this analysis demonstrates that \ours natively integrates effective re-ranking capabilities.

\section{Qualitative Results}
\begin{figure*}[t]
\begin{minipage}[b]{0.325\linewidth}
\scriptsize{\textbf{Q}: What was designated grade II on the same day as\\this building?\vspace{0.018cm}}\\
\begin{minipage}{0.443\linewidth}
\includegraphics[width=1.\linewidth]{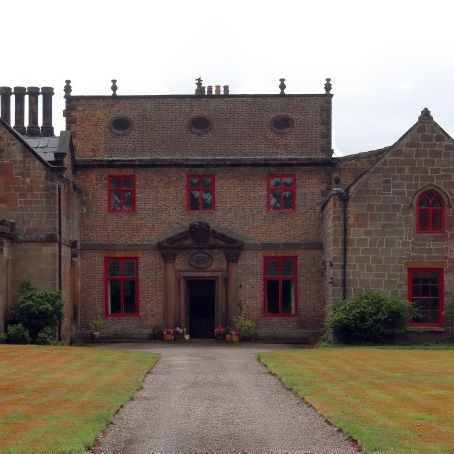}
\end{minipage}
\hfill
\begin{minipage}{0.53\linewidth}
\scriptsize{
\textbf{Wiki-LLaVA~\cite{caffagni2024wiki}}:\\
A of the heritage \textcolor{red}{\xmark} \\
\textbf{EchoSight~\cite{yan2024echosight}}:\\
It seems you are asking about the Grade II listing of the building \textcolor{red}{\xmark} \\
\textbf{\ours (Ours):}\\
Gate Piers \textcolor[HTML]{00b050}{\cmark}
}
\end{minipage}
\end{minipage}
\hspace{0.02cm}
\begin{minipage}[b]{0.325\linewidth}
\scriptsize{\textbf{Q}: What portion of the exterior of this railway station is on Belmont street?\vspace{0.05cm}}\\
\begin{minipage}{0.443\linewidth}
\includegraphics[width=1.\linewidth]{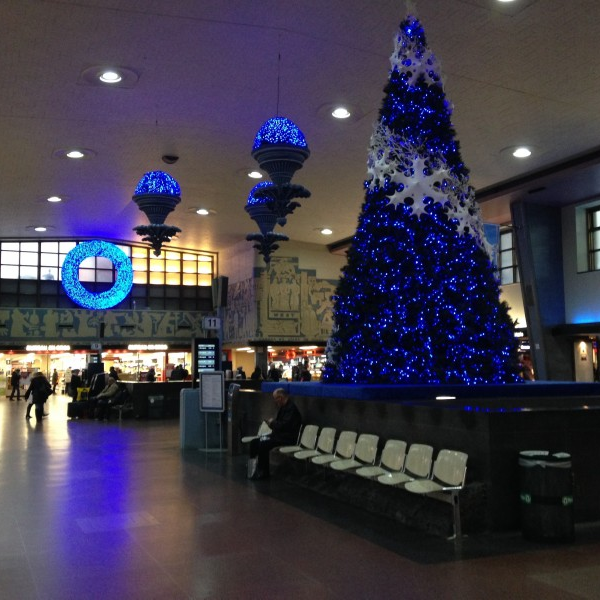}
\end{minipage}
\hfill
\begin{minipage}{0.53\linewidth}
\scriptsize{
\textbf{Wiki-LLaVA~\cite{caffagni2024wiki}}:\\
North \textcolor{red}{\xmark} \\
\textbf{EchoSight~\cite{yan2024echosight}}:\\
I can not verify information about the exterior of the railway station \textcolor{red}{\xmark} \\
\textbf{\ours (Ours):}\\
Only visible portion \textcolor[HTML]{00b050}{\cmark}
}
\end{minipage}
\end{minipage}
\hspace{0.02cm}
\begin{minipage}[b]{0.325\linewidth}
\scriptsize{\textbf{Q}: When was the current structure of this church dedicated?\vspace{0.05cm}}\\
\begin{minipage}{0.443\linewidth}
\includegraphics[width=1.\linewidth]{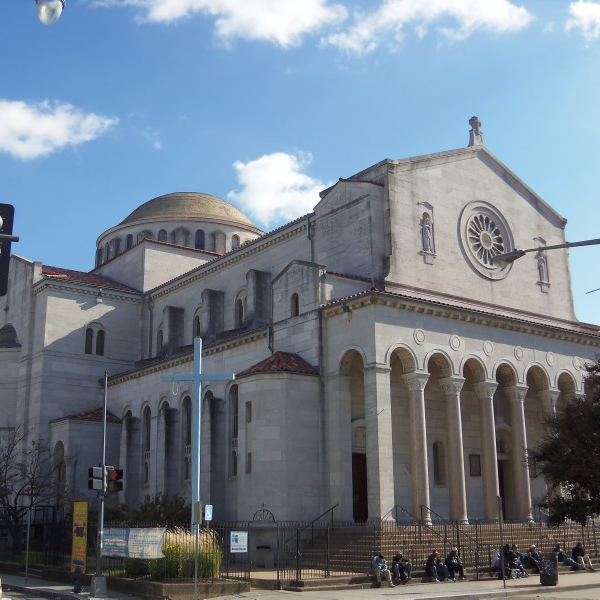}
\end{minipage}
\hfill
\begin{minipage}{0.53\linewidth}
\scriptsize{
\textbf{Wiki-LLaVA~\cite{caffagni2024wiki}}:\\
1931 \textcolor{red}{\xmark} \\
\textbf{EchoSight~\cite{yan2024echosight}}:\\
The provided text does not mention the dedication date of the current structure \textcolor{red}{\xmark} \\
\textbf{\ours (Ours):}\\
1922 \textcolor[HTML]{00b050}{\cmark}
}
\end{minipage}
\end{minipage}
\vspace{0.08cm}

\begin{minipage}[b]{0.325\linewidth}
\scriptsize{\textbf{Q}: Is Fr. Emilio Biosca Agüero the current or former pastor of this church?\vspace{0.11cm}}\\
\begin{minipage}{0.443\linewidth}
\includegraphics[width=1.\linewidth]{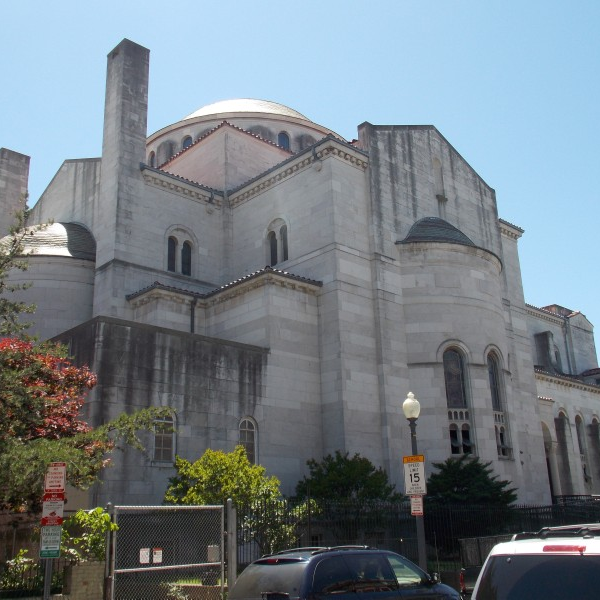}
\end{minipage}
\hfill
\begin{minipage}{0.53\linewidth}
\scriptsize{
\textbf{Wiki-LLaVA~\cite{caffagni2024wiki}}:\\
Former \textcolor{red}{\xmark} \\
\textbf{EchoSight~\cite{yan2024echosight}}:\\
I could not find any information about Fr. Emilio Biosca Agüero \textcolor{red}{\xmark} \\
\textbf{\ours (Ours):}\\
Current \textcolor[HTML]{00b050}{\cmark}
}
\end{minipage}
\end{minipage}
\hspace{0.02cm}
\begin{minipage}[b]{0.325\linewidth}
\scriptsize{\textbf{Q}: What does the oily plumage of this bird prevent its feathers from getting?\vspace{0.06cm}}\\
\begin{minipage}{0.443\linewidth}
\includegraphics[width=1.\linewidth]{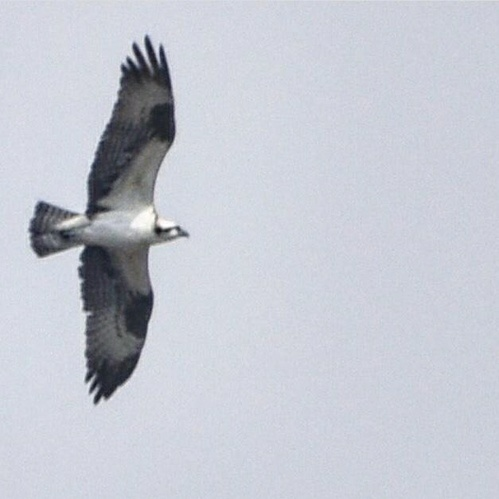}
\end{minipage}
\hfill
\begin{minipage}{0.53\linewidth}
\scriptsize{
\textbf{Wiki-LLaVA~\cite{caffagni2024wiki}}:\\
Wet \textcolor{red}{\xmark} \\
\textbf{EchoSight~\cite{yan2024echosight}}:\\
Unfortunately, the text does not mention a specific bird with oily plumage \textcolor{red}{\xmark} \\
\textbf{\ours (Ours):}\\
Waterlogged \textcolor[HTML]{00b050}{\cmark}
}
\end{minipage}
\end{minipage}
\hspace{0.02cm}
\begin{minipage}[b]{0.325\linewidth}
\scriptsize{\textbf{Q}: What was attached to the western end of the original portion of this building?\vspace{0.05cm}}\\
\begin{minipage}{0.443\linewidth}
\includegraphics[width=1.\linewidth]{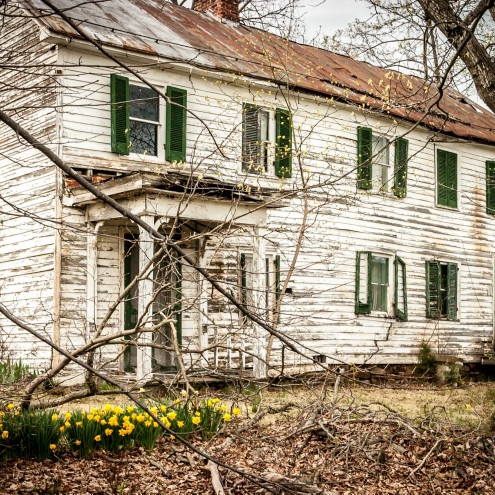}
\end{minipage}
\hfill
\begin{minipage}{0.53\linewidth}
\scriptsize{
\textbf{Wiki-LLaVA~\cite{caffagni2024wiki}}:\\
The United States \textcolor{red}{\xmark} \\
\textbf{EchoSight~\cite{yan2024echosight}}:\\
A kitchen (attached to the house) and living quarters for maids \textcolor{red}{\xmark} \\
\textbf{\ours (Ours):}\\
Chimney \textcolor[HTML]{00b050}{\cmark}
}
\end{minipage}
\end{minipage}
\vspace{0.08cm}

\begin{minipage}[b]{0.325\linewidth}
\scriptsize{\textbf{Q}: How did this bird fare against the Australian white ibis?\vspace{0.05cm}}\\
\begin{minipage}{0.443\linewidth}
\includegraphics[width=1.\linewidth]{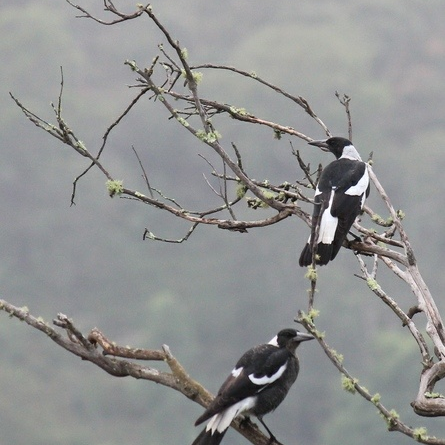}
\end{minipage}
\hfill
\begin{minipage}{0.53\linewidth}
\scriptsize{
\textbf{Wiki-LLaVA~\cite{caffagni2024wiki}}:\\
The world's largest \textcolor{red}{\xmark} \\
\textbf{EchoSight~\cite{yan2024echosight}}:\\
There is no mention of the Australian white ibis in the provided text \textcolor{red}{\xmark} \\
\textbf{\ours (Ours):}\\
Narrowly ahead \textcolor[HTML]{00b050}{\cmark}
}
\end{minipage}
\end{minipage}
\hspace{0.02cm}
\begin{minipage}[b]{0.325\linewidth}
\scriptsize{\textbf{Q}: How often is the kite festival held at this castle?\vspace{0.05cm}}\\
\begin{minipage}{0.443\linewidth}
\includegraphics[width=1.\linewidth]{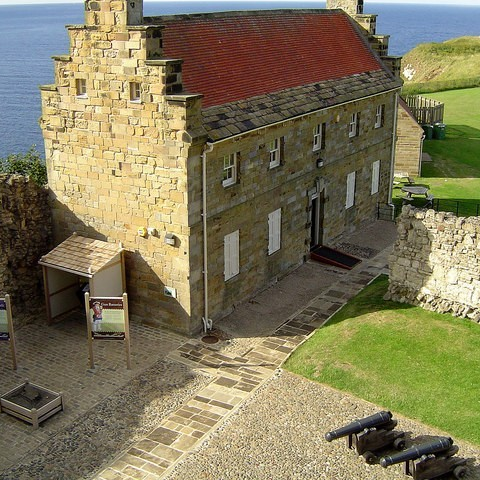}
\end{minipage}
\hfill
\begin{minipage}{0.53\linewidth}
\scriptsize{
\textbf{Wiki-LLaVA~\cite{caffagni2024wiki}}:\\
Never \textcolor{red}{\xmark} \\
\textbf{EchoSight~\cite{yan2024echosight}}:\\
There is no information in the provided text about the frequency of a kite festival \textcolor{red}{\xmark} \\
\textbf{\ours (Ours):}\\
Annual \textcolor[HTML]{00b050}{\cmark}
}
\end{minipage}
\end{minipage}
\hspace{0.02cm}
\begin{minipage}[b]{0.325\linewidth}
\scriptsize{\textbf{Q}: What kind of insect is this?\vspace{0.05cm}}\\
\begin{minipage}{0.443\linewidth}
\includegraphics[width=1.\linewidth]{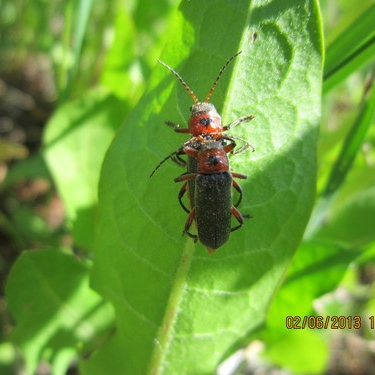}
\end{minipage}
\hfill
\begin{minipage}{0.53\linewidth}
\scriptsize{
\textbf{Wiki-LLaVA~\cite{caffagni2024wiki}}:\\
Aa wasp, the spider \textcolor{red}{\xmark} \\
\textbf{EchoSight~\cite{yan2024echosight}}:\\
The Watercress Darter is actually a type of fish, not an insect \textcolor{red}{\xmark} \\
\textbf{\ours (Ours):}\\
Beetles \textcolor[HTML]{00b050}{\cmark}
}
\end{minipage}
\end{minipage}
\vspace{-0.2cm}
\caption{Additional qualitative results on image-question pairs from Encyclopedic-VQA, where we compare the answers provided by \ours with those from WikiLLaVA~\cite{caffagni2024wiki} and EchoSight~\cite{yan2024echosight}.}
\label{fig:qualitatives_supp1}
\vspace{-0.15cm}
\end{figure*}

\begin{figure*}[t]
\begin{minipage}[b]{0.325\linewidth}
\scriptsize{\textbf{Q}: What is the area in square kilometer occupied by \\this lake?\vspace{0.11cm}}\\
\begin{minipage}{0.443\linewidth}
\includegraphics[width=1.\linewidth]{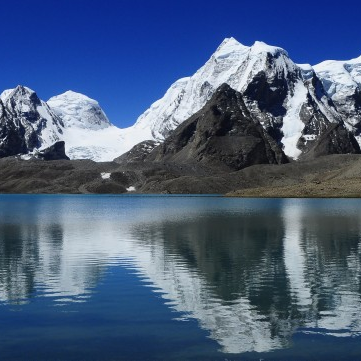}
\end{minipage}
\hfill
\begin{minipage}{0.53\linewidth}
\scriptsize{
\textbf{Wiki-LLaVA~\cite{caffagni2024wiki}}:\\
9.82 \textcolor{red}{\xmark} \\
\textbf{EchoSight~\cite{yan2024echosight}}:\\
5.34 \textcolor{red}{\xmark} \\
\textbf{\ours (Ours):}\\
1.18 \textcolor[HTML]{00b050}{\cmark}
}
\end{minipage}
\end{minipage}
\hspace{0.02cm}
\begin{minipage}[b]{0.325\linewidth}
\scriptsize{\textbf{Q}: Which class of biological feature is this food\\ produced by?\vspace{0.05cm}}\\
\begin{minipage}{0.443\linewidth}
\includegraphics[width=1.\linewidth]{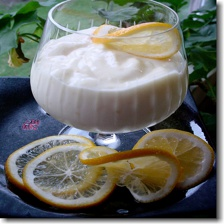}
\end{minipage}
\hfill
\begin{minipage}{0.53\linewidth}
\scriptsize{
\textbf{Wiki-LLaVA~\cite{caffagni2024wiki}}:\\
Malt house \textcolor{red}{\xmark} \\
\textbf{EchoSight~\cite{yan2024echosight}}:\\
Plants \textcolor{red}{\xmark} \\
\textbf{\ours (Ours):}\\
Lactobacillus delbrueckii \textcolor[HTML]{00b050}{\cmark}
}
\end{minipage}
\end{minipage}
\hspace{0.02cm}
\begin{minipage}[b]{0.325\linewidth}
\scriptsize{\textbf{Q:} What is the density (in gram per cubic centimeter)\\ of this place?\vspace{0.05cm}}\\
\begin{minipage}{0.443\linewidth}
\includegraphics[width=1.\linewidth]{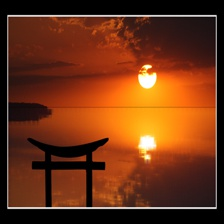}
\end{minipage}
\hfill
\begin{minipage}{0.53\linewidth}
\scriptsize{
\textbf{Wiki-LLaVA~\cite{caffagni2024wiki}}:\\
100 \textcolor{red}{\xmark} \\
\textbf{EchoSight~\cite{yan2024echosight}}:\\
There is no information about the density of this place \textcolor{red}{\xmark} \\
\textbf{\ours (Ours):}\\
1408 \textcolor[HTML]{00b050}{\cmark}
}
\end{minipage}
\end{minipage}
\vspace{0.08cm}

\begin{minipage}[b]{0.325\linewidth}
\scriptsize{\textbf{Q}: What is the source of energy of this vehicle?\vspace{0.05cm}}\\
\begin{minipage}{0.443\linewidth}
\includegraphics[width=1.\linewidth]{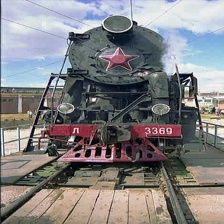}
\end{minipage}
\hfill
\begin{minipage}{0.53\linewidth}
\scriptsize{
\textbf{Wiki-LLaVA~\cite{caffagni2024wiki}}:\\
Diesel \textcolor{red}{\xmark} \\
\textbf{EchoSight~\cite{yan2024echosight}}:\\
Diesel \textcolor{red}{\xmark} \\
\textbf{\ours (Ours):}\\
Coal \textcolor[HTML]{00b050}{\cmark}
}
\end{minipage}
\end{minipage}
\hspace{0.02cm}
\begin{minipage}[b]{0.325\linewidth}
\scriptsize{\textbf{Q}: Where are organization initially formed?\vspace{0.05cm}}\\
\begin{minipage}{0.443\linewidth}
\includegraphics[width=1.\linewidth]{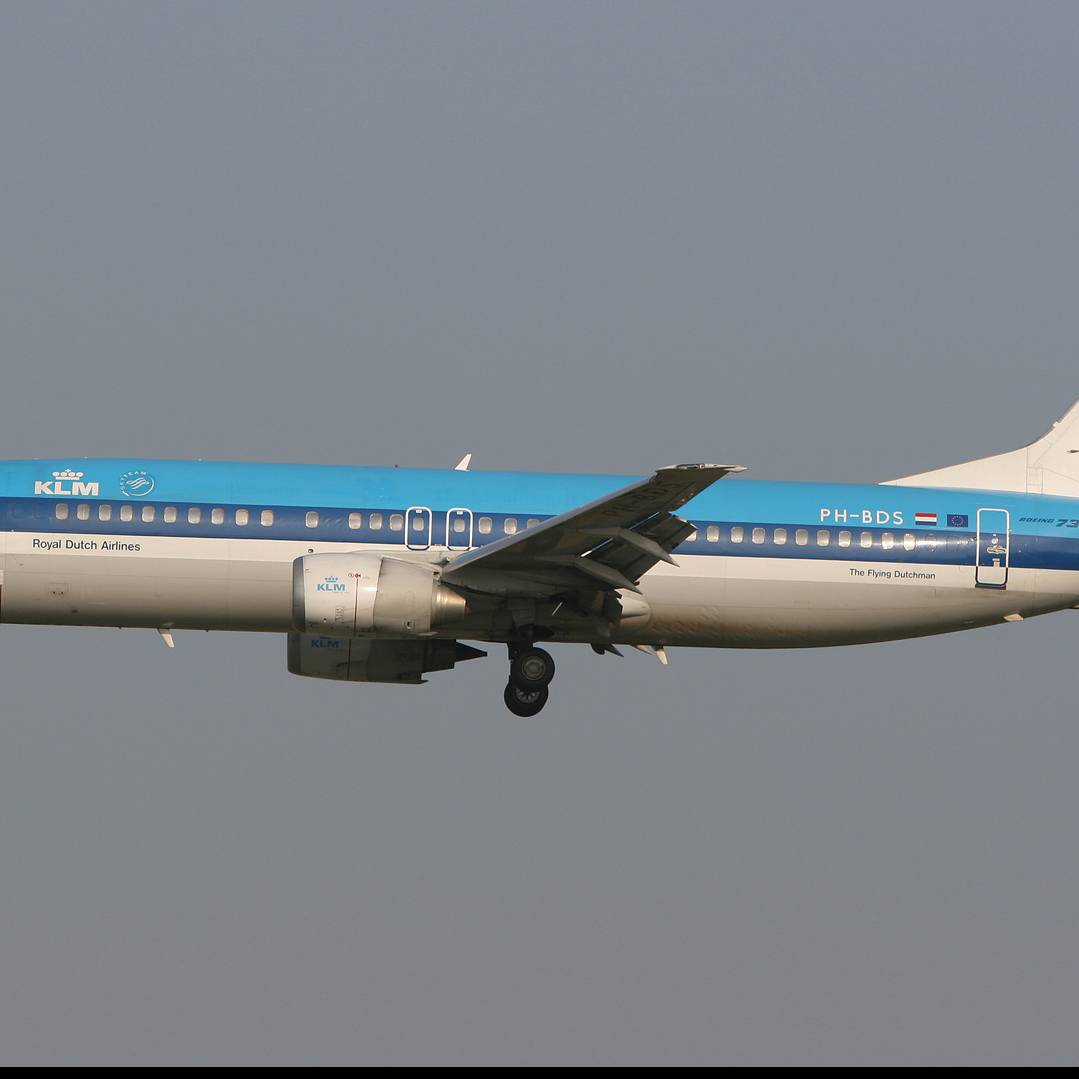}
\end{minipage}
\hfill
\begin{minipage}{0.53\linewidth}
\scriptsize{
\textbf{Wiki-LLaVA~\cite{caffagni2024wiki}}:\\
Wright brother's brother \textcolor{red}{\xmark} \\
\textbf{EchoSight~\cite{yan2024echosight}}:\\
I cannot provide information or context about the accident of Transbrasil Flight 801 \textcolor{red}{\xmark} \\
\textbf{\ours (Ours):}\\
Europe \textcolor[HTML]{00b050}{\cmark}
}
\end{minipage}
\end{minipage}
\hspace{0.02cm}
\begin{minipage}[b]{0.325\linewidth}
\scriptsize{\textbf{Q}: Which place is this animal endemic to?\vspace{0.05cm}}\\
\begin{minipage}{0.443\linewidth}
\includegraphics[width=1.\linewidth]{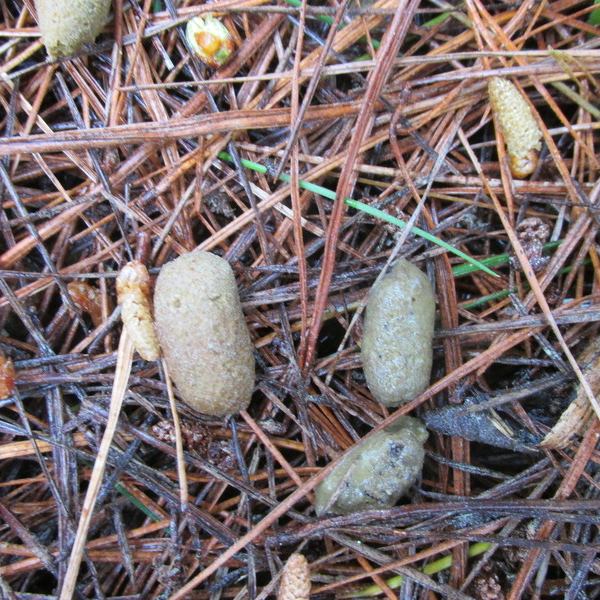}
\end{minipage}
\hfill
\begin{minipage}{0.53\linewidth}
\scriptsize{
\textbf{Wiki-LLaVA~\cite{caffagni2024wiki}}:\\
North Pacific Ocean \textcolor{red}{\xmark} \\
\textbf{EchoSight~\cite{yan2024echosight}}:\\
North America \textcolor{red}{\xmark} \\
\textbf{\ours (Ours):}\\
New Zealand \textcolor[HTML]{00b050}{\cmark}
}
\end{minipage}
\end{minipage}
\vspace{0.08cm}

\begin{minipage}[b]{0.325\linewidth}
\scriptsize{\textbf{Q}: What is the architectural style of this place?\vspace{0.05cm}}\\
\begin{minipage}{0.443\linewidth}
\includegraphics[width=1.\linewidth]{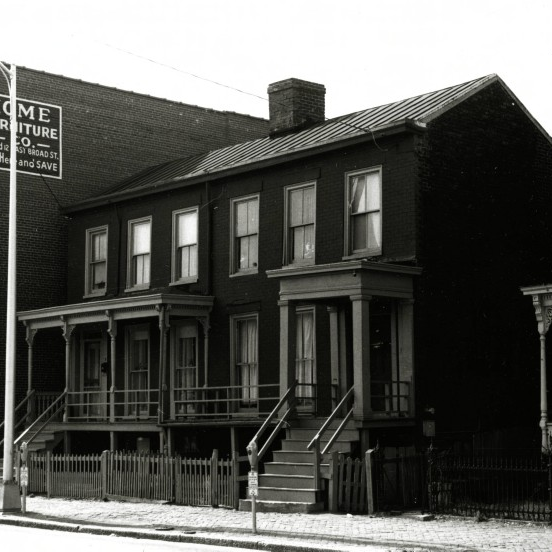}
\end{minipage}
\hfill
\begin{minipage}{0.53\linewidth}
\scriptsize{
\textbf{Wiki-LLaVA~\cite{caffagni2024wiki}}:\\
There is no specific answer to the question about the architectural style in the text \textcolor{red}{\xmark} \\
\textbf{EchoSight~\cite{yan2024echosight}}:\\
Georgian architecture \textcolor{red}{\xmark} \\
\textbf{\ours (Ours):}\\
Greek Revival architecture \textcolor[HTML]{00b050}{\cmark}
}
\end{minipage}
\end{minipage}
\hspace{0.02cm}
\begin{minipage}[b]{0.325\linewidth}
\scriptsize{\textbf{Q}: Which crystal system does this material have?\vspace{0.05cm}}\\
\begin{minipage}{0.443\linewidth}
\includegraphics[width=1.\linewidth]{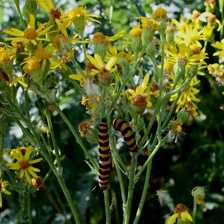}
\end{minipage}
\hfill
\begin{minipage}{0.53\linewidth}
\scriptsize{
\textbf{Wiki-LLaVA~\cite{caffagni2024wiki}}:\\
Hexagonal \textcolor{red}{\xmark} \\
\textbf{EchoSight~\cite{yan2024echosight}}:\\
There is no crystal system mentioned in the text, so I will say: None \textcolor{red}{\xmark} \\
\textbf{\ours (Ours):}\\
Trigonal \textcolor[HTML]{00b050}{\cmark}
}
\end{minipage}
\end{minipage}
\hspace{0.02cm}
\begin{minipage}[b]{0.325\linewidth}
\scriptsize{\textbf{Q:} Which street is this building located at?\vspace{0.05cm}}\\
\begin{minipage}{0.443\linewidth}
\includegraphics[width=1.\linewidth]{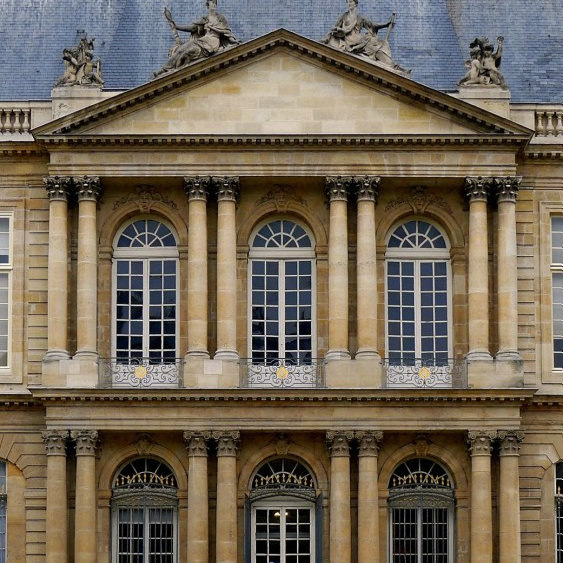}
\end{minipage}
\hfill
\begin{minipage}{0.53\linewidth}
\scriptsize{
\textbf{Wiki-LLaVA~\cite{caffagni2024wiki}}:\\
Rue de Rivoli \textcolor{red}{\xmark} \\
\textbf{EchoSight~\cite{yan2024echosight}}:\\
There is no street mentioned in the text \textcolor{red}{\xmark} \\
\textbf{\ours (Ours):}\\
Rue des Francs-Bourgeois \textcolor[HTML]{00b050}{\cmark}
}
\end{minipage}
\end{minipage}
\vspace{-0.2cm}
\caption{Additional qualitative results on image-question pairs from InfoSeek, where we compare the answers provided by \ours with those from WikiLLaVA~\cite{caffagni2024wiki} and EchoSight~\cite{yan2024echosight}.}
\label{fig:qualitatives_supp2}
\vspace{-0.3cm}
\end{figure*}

To comprehensively evaluate the proposed \ours model, we present additional qualitative results in Fig.~\ref{fig:qualitatives_supp1} and Fig.~\ref{fig:qualitatives_supp2}, alongside answers generated by Wiki-LLaVA~\cite{caffagni2024wiki} and EchoSight~\cite{yan2024echosight}. These results are based on sample image-question pairs from Encyclopedic-VQA and InfoSeek, respectively. Notably, \ours effectively handles diverse question types, including those requiring numeric answers such as dates or measurements (\eg~the top-right example in Fig.~\ref{fig:qualitatives_supp1} and the top-left example in Fig.~\ref{fig:qualitatives_supp2}), as well as questions demanding domain-specific knowledge (\eg~the top-left example in Fig.~\ref{fig:qualitatives_supp1} and the bottom-left example in Fig.~\ref{fig:qualitatives_supp2}). Conversely, both Wiki-LLaVA and EchoSight often fail to answer the given questions correctly, either generating an incorrect response or a statement indicating their inability to provide an answer. These results qualitatively highlight the superior performance of \ours compared to existing methods for the task.

\begin{table}[t]
  \centering
  \setlength{\tabcolsep}{.42em}
  \resizebox{0.75\linewidth}{!}{
  \begin{tabular}{cccc cc}
   \toprule
    & & & & \multicolumn{2}{c}{\textbf{E-VQA}} \\
    \cmidrule{5-6}
     \textbf{Built-In Re-Ranking} & $k$ & $k_p$ & & Single-Hop & All \\
    \midrule
    \xmark & 5 & - & & \textbf{35.5} & \textbf{35.5} \\
    \cmark & 5 & 1 & & 34.7 & 34.8 \\
    \cmark & 5 & 2 & & 35.0 & 34.9 \\
    \cmark & 5 & 5 & & 33.4 & 33.4 \\
    \midrule
    \xmark & 20 & - & & 35.7 & 35.2 \\
    \cmark & 20 & 1 & & 36.3 & 35.8 \\
    \cmark & 20 & 2 & & \textbf{36.6} & \textbf{36.6} \\
    \cmark & 20 & 5 & & 35.9 & 35.6 \\
    \midrule
    \xmark & 50 & - & & 29.6 & 29.1 \\
    \cmark & 50 & 1 & & 37.6 & 37.1 \\
    \cmark & 50 & 2 & & \underline{\textbf{37.8}} & \underline{\textbf{37.2}} \\
    \cmark & 50 & 5 & & 36.7 & 36.3 \\
  \bottomrule
  \end{tabular}
  }
  \vspace{-0.2cm}
\caption{Experimental analysis of built-in re-ranking capabilities on the Encyclopedic-VQA test set, varying the number $k$ of retrieved documents and the number $k_p$ of selected passages after re-ranking. Best results for each $k$ are in bold, and the best overall results are underlined.}
\label{tab:reranking_probs}
\vspace{-0.35cm}
\end{table}

\section{Limitations and Failure Cases}
\begin{figure*}[t]
\begin{minipage}[b]{0.325\linewidth}
\scriptsize{\textbf{Q}: Terry park became the pirates spring what ballpark?\vspace{0.05cm}}\\
\begin{minipage}{0.443\linewidth}
\includegraphics[width=1.\linewidth]{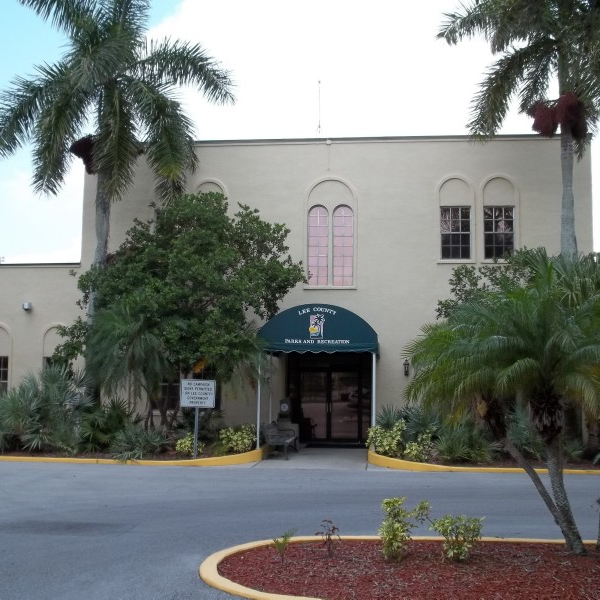}
\end{minipage}
\hfill
\begin{minipage}{0.53\linewidth}
\scriptsize{
\textbf{Ground-truth}:\\
Training home\\\\
\textbf{\ours (Ours):}\\
Training ground
}
\end{minipage}
\end{minipage}
\hspace{0.02cm}
\begin{minipage}[b]{0.325\linewidth}
\scriptsize{\textbf{Q}: In what cuisine is this vegetable used?\vspace{0.05cm}}\\
\begin{minipage}{0.443\linewidth}
\includegraphics[width=1.\linewidth]{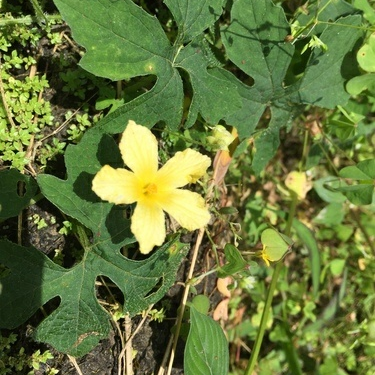}
\end{minipage}
\hfill
\begin{minipage}{0.53\linewidth}
\scriptsize{
\textbf{Ground-truth}:\\
Southeast Asia\\\\
\textbf{\ours (Ours):}\\
Indian cuisine
}
\end{minipage}
\end{minipage}
\hspace{0.02cm}
\begin{minipage}[b]{0.325\linewidth}
\scriptsize{\textbf{Q:} What is the fee to see this gate from the outside?\vspace{0.05cm}}\\
\begin{minipage}{0.443\linewidth}
\includegraphics[width=1.\linewidth]{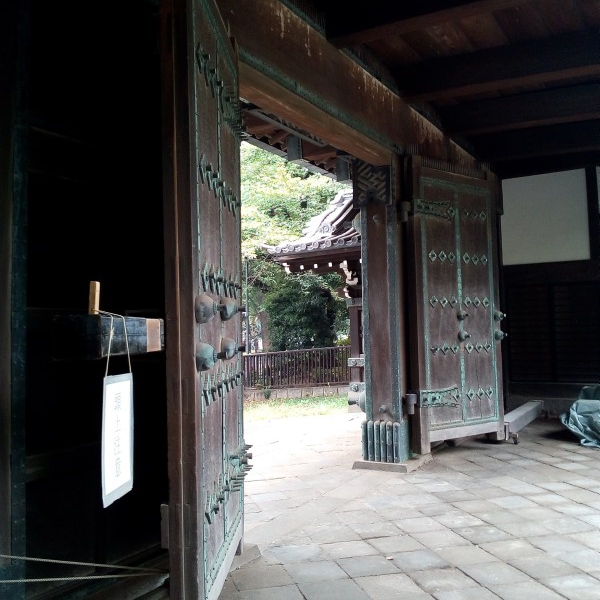}
\end{minipage}
\hfill
\begin{minipage}{0.53\linewidth}
\scriptsize{
\textbf{Ground-truth}:\\
No admission fee \\\\
\textbf{\ours (Ours):}\\
Free
}
\end{minipage}
\end{minipage}
\vspace{0.08cm}

\begin{minipage}[b]{0.325\linewidth}
\scriptsize{\textbf{Q}: What is the objective of this object?\vspace{0.05cm}}\\
\begin{minipage}{0.443\linewidth}
\includegraphics[width=1.\linewidth]{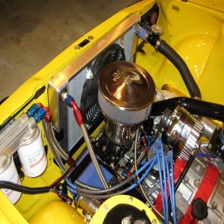}
\end{minipage}
\hfill
\begin{minipage}{0.53\linewidth}
\scriptsize{
\textbf{Ground-truth}:\\
Energy transformation\\\\
\textbf{\ours (Ours):}\\
Engine tuning
}
\end{minipage}
\end{minipage}
\hspace{0.02cm}
\begin{minipage}[b]{0.325\linewidth}
\scriptsize{\textbf{Q}: What is the magnitude of the velocity of this\\ animal kilometer per hour?\vspace{0.05cm}}\\
\begin{minipage}{0.443\linewidth}
\includegraphics[width=1.\linewidth]{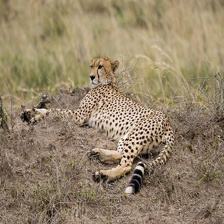}
\end{minipage}
\hfill
\begin{minipage}{0.53\linewidth}
\scriptsize{
\textbf{Ground-truth}:\\
120\\\\
\textbf{\ours (Ours):}\\
150
}
\end{minipage}
\end{minipage}
\hspace{0.02cm}
\begin{minipage}[b]{0.325\linewidth}
\scriptsize{\textbf{\\Q:} What is the conservation status of this bird?\vspace{0.05cm}}\\
\begin{minipage}{0.443\linewidth}
\includegraphics[width=1.\linewidth]{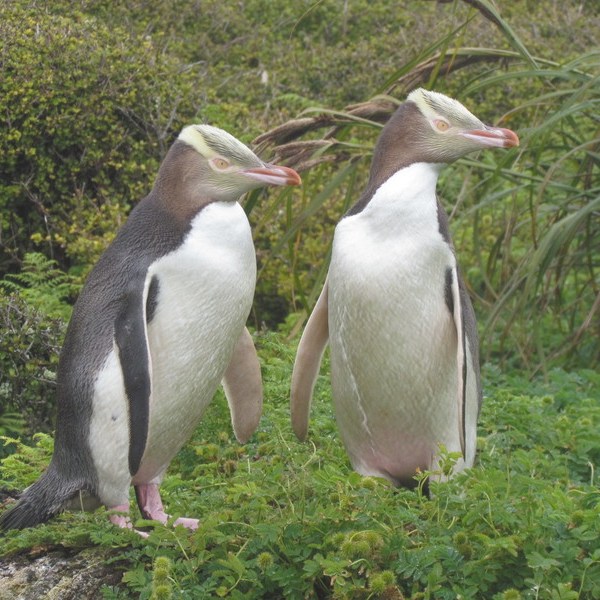}
\end{minipage}
\hfill
\begin{minipage}{0.53\linewidth}
\scriptsize{
\textbf{Ground-truth}:\\
Endangered\\\\
\textbf{\ours (Ours):}\\
Vulnerable
}
\end{minipage}
\end{minipage}
\vspace{-0.2cm}
\caption{Examples of failure cases on image-question pairs from Encyclopedic-VQA (top row) and InfoSeek (bottom row).}
\label{fig:failure_cases}
\vspace{-0.2cm}
\end{figure*}

As the final part of the qualitative analysis, we present some failure cases of our model in Fig.~\ref{fig:failure_cases}. These examples illustrate the challenges the model faces, such as adhering to the expected style of correct answers. This issue is particularly evident in datasets like InfoSeek, where the evaluation is based on exact matches between ground-truth and generated answers. This is evident in the bottom-left example, where the answer conveys the same semantic meaning but is expressed in a different way. This may lead to discrepancies when evaluating the correctness of the given answer despite its semantic alignment with the ground-truth. Other errors arise from the specificity of the questions. For instance, in the bottom-center example, the question asks about the maximum velocity of a cheetah, requiring high precision. In this case, \ours provides a reasonable estimate, demonstrating its ability to address such detailed inquiries.

\end{document}